\pdfoutput=1

\documentclass[11pt]{article}

\usepackage{acl}

\usepackage[T1]{fontenc}
\usepackage[utf8]{inputenc}
\usepackage{times}
\usepackage{latexsym}
\usepackage{microtype}
\usepackage{inconsolata}
\usepackage{graphicx}
\usepackage{booktabs}
\usepackage{amsmath}
\usepackage{amssymb}
\usepackage{amsfonts}
\usepackage{nicefrac}
\usepackage{multirow}
\usepackage{xcolor}
\usepackage{enumitem}
\usepackage{subcaption}
\usepackage{url}
\usepackage{tikz}
\usetikzlibrary{arrows.meta, positioning, shapes.geometric, fit, backgrounds, calc}
\graphicspath{{figures/}}

\title{RAW: Refusal Aversion in the Wild,\\A Causal Measurement Method for Deployed LLMs}

\author{Pavel Kireyev \\
  Department of Management \\
  London School of Economics and Political Science \\
  \texttt{p.kireyev@lse.ac.uk}}

\hypersetup{pdftitle={RAW: Refusal Aversion in the Wild, A Causal Measurement Method for Deployed LLMs},
            pdfauthor={Pavel Kireyev}}

\begin{document}
\maketitle

\begin{figure*}[t!]
\centering
\begin{tikzpicture}[
  >=Latex,
  font=\small,
  pipestep/.style={rectangle, draw=blue!50!black, rounded corners=3pt,
               minimum width=3.35cm, minimum height=1.6cm, align=center,
               fill=blue!6, thick, line width=0.5pt,
               inner sep=4pt},
  artifact/.style={rectangle, draw=gray!50, rounded corners=2pt,
                   minimum width=3.35cm, minimum height=1.7cm, align=center,
                   fill=gray!4, font=\scriptsize, line width=0.4pt,
                   inner sep=5pt},
  bigarrow/.style={->, line width=1.0pt, draw=blue!50!black},
  thinarrow/.style={->, line width=0.5pt, draw=gray!60},
]
\node[pipestep] (s1) at (0, 0)    {\textbf{1.\ Conversation log}\\[1pt]
  \footnotesize Existing log from\\\footnotesize a deployed LLM};
\node[pipestep] (s2) at (4.2, 0)  {\textbf{2.\ Label \& encode}\\[1pt]
  \footnotesize $T$: refusal label (judge)\\\footnotesize $X{=}$embedding $+$ score\\\footnotesize $Y$: turn-3 reply};
\node[pipestep] (s3) at (8.4, 0)  {\textbf{3.\ Match}\\[1pt]
  \footnotesize Near-duplicate clusters\\\footnotesize $\cos\geq 0.7$,\\\footnotesize matched $X_s$};
\node[pipestep] (s4) at (12.6, 0) {\textbf{4.\ ATT via AIPW}\\[1pt]
  \footnotesize Cross-fit, doubly robust\\\footnotesize $+$ 3 secondary\\\footnotesize estimators};
\draw[bigarrow] (s1) -- (s2);
\draw[bigarrow] (s2) -- (s3);
\draw[bigarrow] (s3) -- (s4);

\node[artifact, below=0.55cm of s1] (a1) {%
  \emph{User:} ``how do I bypass\\ the API rate limit?''\\[1pt]
  \emph{LLM:} (varies by sampler)\\[1pt]
  \emph{Next turn?} yes/no};
\node[artifact, below=0.55cm of s2] (a2) {%
  $T_i \in \{0,1\}$\\[1pt]
  $X_i \in \mathbb{R}^{1025}$\\[1pt]
  $Y_i \in \{0,1\}$};
\node[artifact, below=0.55cm of s3] (a3) {%
  $P_a\colon$ {\color{red!70!black}refuse}, $Y{=}0$\\[1pt]
  $P_b\colon$ {\color{teal!60!black}comply}, $Y{=}1$\\[1pt]
  $P_c\colon$ {\color{teal!60!black}comply}, $Y{=}0$\\[2pt]
  $T$ varies $\Rightarrow$ SoO};
\node[artifact, below=0.55cm of s4] (a4) {%
  \emph{Koala-13b LMSYS:}\\[1pt]
  unadj.\ $-21$\,pp\\
  AIPW $-3$\,pp\\[1pt]
  $\Rightarrow$ $18$\,pp prompt bias};
\draw[thinarrow] (s1) -- (a1);
\draw[thinarrow] (s2) -- (a2);
\draw[thinarrow] (s3) -- (a3);
\draw[thinarrow] (s4) -- (a4);

\end{tikzpicture}
\caption{The RAW pipeline. (1)~A conversation log from any deployed LLM. (2)~Each conversation is encoded into a refusal label $T$ (LLM-as-judge), a $1025$-dim prompt-content covariate $X$ (\texttt{mxbai-embed-large} embedding concatenated with the OpenAI Moderation \texttt{max\_concern\_score} $X_s$), and a re-engagement outcome $Y$ (user sends turn~3). (3)~Within matched clusters of near-duplicate prompts the same model produces both refusal and compliance at near-fixed $X$ (three matched conversations are shown: $P_a$ refuses and the user does not re-engage, $P_b$ complies and the user re-engages, $P_c$ complies and the user does not), so the residual variation in $T$ is sampler-driven, satisfying selection on observables (SoO). (4)~Cross-fit augmented inverse-propensity weighting (AIPW) averages per-conversation influence-function contributions to recover the average treatment effect on the treated (ATT), with three secondary estimators (score, embedding+score, Crump-trimmed AIPW) as cross-checks.}
\label{fig:method}
\end{figure*}
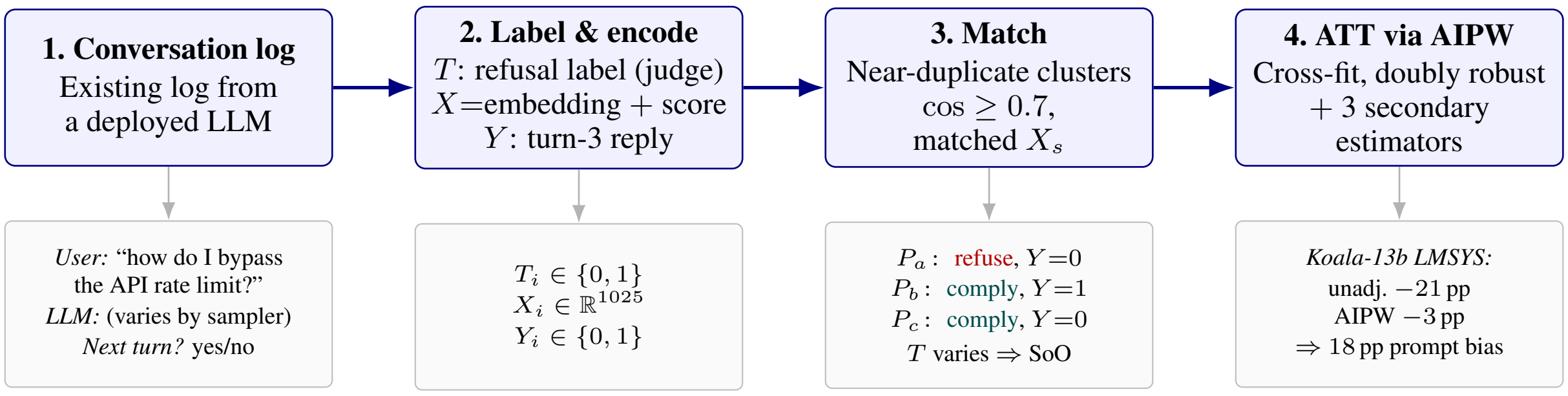

\begin{abstract}
Deployed LLMs field hundreds of millions of conversations a day and refuse a non-trivial fraction of user requests, yet the user-side cost of refusal, whether users disengage and by how much, is unmeasured on any production model: a blind spot for the consumer-facing platforms built on them. A/B-testing refusal is ethically and operationally constrained on frontier-lab products and unavailable to third-party platforms deploying these models through a fixed API. We introduce RAW, a method that uses LLM stochasticity as a natural experiment: at near-identical prompts in any existing log, the same model sometimes refuses and sometimes does not, and that variation identifies the causal user-side effect of refusal. Applying the method to 15 deployed LLMs in LMSYS-Chat-1M and WildChat, with user-clustered inference, we find that for about half of the models we can measure precisely, a refusal costs 3 to 17 pp of subsequent re-engagement; the rest show no distinguishable effect, among them one deployment whose positive point estimate we trace to persistent users retrying the refused request. Our contribution is the measurement method, which runs on any deployed model from logs alone for about \$0.30 per 1,000 conversations. We release labels, moderation scores, prompt embeddings, and pipeline code.
\end{abstract}

\section{Introduction}
\label{sec:intro}

Large language models are deployed into hundreds of millions of conversations a day. In a non-trivial fraction of them, the model declines the user's request, citing safety, policy, or capability. Over-refusal is itself a named harm in the model-evaluation literature \cite{rottger2024xstest,cui2024orbench}. What remains unmeasured at scale is the user-side consequence: when a refusal occurs in deployed chat, do users disengage, and by how much? We call this consequence \emph{refusal aversion}, the tendency of users to disengage after being refused; it is a property of how a model's users react to a refusal, not of the model's own willingness to refuse.

This gap matters most to the consumer-facing products built on LLMs: conversational and companion apps, role-play and character platforms, customer-service bots, and content tools with chat features. These platforms reach the underlying models only through a fixed provider API, observe their own users' conversation logs, and are precisely where users submit toxic, sensitive, or otherwise risky content and meet refusals. For them a refusal is not an abstract safety event but a churn driver: a safety-tuned model trades user retention for harm reduction, the helpful-versus-harmless tradeoff at the heart of alignment \cite{bai2022hh}, yet no operator can measure how much retention their guardrails cost. This cost surfaces as within-session drop-off, one component of churn and user lifetime value; a single chatbot non-progress error precedes ${\approx}45\%$ of long-term abandonments, and three or fewer precede $90\%$ \cite{li2020noprogress}. RAW measures this within-conversation cost, not longer-horizon or account-level churn. Crucially, these operators cannot A/B-test the underlying model's refusal behavior: switching providers entangles refusal with capability, style, latency, and cost at once (Section~\ref{sec:why_not_ab}).

Our method turns one fact about deployed models into a measurement of refusal aversion. By a \emph{deployed LLM} we mean a fixed model alias serving real user traffic, whose conversation logs the analyst holds, and whose responses are sampled at non-zero temperature; Section~\ref{sec:method} unpacks this definition. LLMs are stochastic: at a fixed prompt, the same model sometimes refuses and sometimes does not. Across near-identical prompts in a conversation log (the same paraphrased question or risky scenario), which session gets refused is decided partly by the sampler, not by anything that separately drives whether the user keeps going. Conditional on prompt content (a sentence embedding) and moderation risk (a scalar score), refusal versus compliance is a coin-flip the operator never had to run, so the effect of a refusal on the user's next action is identified from logs the operator already has. Turning that observation into a credible number, through a matching design, doubly robust estimation, and explicit sensitivity analysis, is this paper's central contribution; Figure~\ref{fig:method} sketches the full pipeline end to end.

\paragraph{Contributions.}
\begin{enumerate}[leftmargin=1.5em,itemsep=2pt]
  \item \emph{A causal measurement method for deployed-LLM user-side effects.} We formalize within-prompt LLM stochasticity as a quasi-randomization device, identify the average treatment effect on the treated (ATT) under selection on observables given prompt content and moderation risk, and estimate it with four complementary estimators backed by overlap and sensitivity diagnostics. The method runs on existing conversation logs, with no experiment or risky-prompt exposure required.
  \item \emph{An empirical demonstration} on conversation logs covering $15$ deployed LLMs in LMSYS-Chat-1M \cite{zheng2024lmsyschat1m} and WildChat \cite{zhao2024wildchat}. We analyze one model on one conversation pool at a time, a unit we call a \emph{cell}; the $15$ models span $19$ cells, since four GPT models are each evaluated on two pools (Section~\ref{sec:corpus}). On the $13$ cells with enough data for a precise estimate, the \emph{main panel}, and with inference clustered on users wherever the source data identify them, RAW finds significant refusal aversion in $7$ cells and no statistically distinguishable effect in $6$, including the only pretraining-only model and one positive point estimate that we attribute to that cell's persistent user population rather than the model. Because costs vary by user population, the relevant number is deployment-specific, not a model-side benchmark lookup, which is what makes a log-based measurement necessary.
  \item \emph{The RAW corpus}: multi-judge refusal labels, moderation scores, and prompt embeddings for the $19$-cell panel, with conversation-hash join keys back to source, released to enable replication on new models and extension to new response attributes (verbosity, hedging, formatting).
\end{enumerate}

\section{Related work}
\label{sec:related}

Refusal benchmarks \cite{rottger2024xstest,cui2024orbench,tedeschi2024alert,ji2023beavertails,chao2024jbb,xie2024sorrybench,mazeika2024harmbench} evaluate \emph{model} behavior given a prompt; they say nothing about how users respond when a model refuses. The closest user-side priors split into two camps. Small controlled studies measure \emph{perceptions} rather than behavior: \citet{wester2024denials} ($N{\approx}105$) and \citet{zheng2025letthemdown} ($N{\approx}480$) both report Likert outcomes. One naturalistic preference study, \citet{pasch2025refusal}, uses ${\approx}50$K votes from Chatbot Arena \citep{zheng2023llmjudge} to show that ethical refusals cut pairwise win rate from $36\%$ to $8\%$, but its outcome is a vote, not continuation, and it makes no causal adjustment. Closest in spirit, \citet{defreitas2025manipulation} audit real farewell messages across popular companion apps and run preregistered experiments showing that emotionally manipulative goodbyes causally increase post-farewell engagement; but their treatment is an app-injected message, established by experiment, not the user-side cost of a refusal recovered from logs. None of this prior work measures the causal effect of a refusal itself, on deployed models, from logs the operator already holds, which is the gap RAW fills.

Methodologically, RAW draws on the text-causal-inference literature \cite{feder2022causal}, whose two main configurations treat text as the treatment, e.g., the causal effect of a linguistic property \cite{pryzant2021causal,egami2022causal}, or adjust for and match on text as the confounder \cite{veitch2020embeddings,roberts2020matching,mozer2020matching}. RAW uses the second configuration with a treatment that literature has not studied: the \emph{LLM's own choice of response} at a fixed prompt, with the prompt text as the confounder and model stochasticity at fixed prompts as the randomization device. Refusal is a clean leading case, binary and labelable at scale, but the same design applies to any response attribute that varies stochastically at a fixed prompt, such as verbosity, hedging, or formatting. Identification builds on \citet{rosenbaum1983propensity}; estimation on augmented inverse-propensity weighting (AIPW; \citealp{robins1994ipw}) and cross-fit double machine learning \cite{chernozhukov2018dml,kennedy2024dml}; sensitivity on Rosenbaum $\Gamma$ \cite{rosenbaum2002observational}; overlap on \citet{crump2009overlap}. Online controlled experimentation \cite{kohavi2020trustworthy} and production LLM-ops tooling supply the prospective A/B alternative we complement (Section~\ref{sec:why_not_ab}). Finally, RAW's treatment variable is itself an LLM-as-judge refusal label \cite{zheng2023llmjudge,panickssery2024self,wataoka2024,chen2024judgebias,ye2024justice}, for which we use a coalesced gpt-5.4-mini judge with an open-weights fallback (Section~\ref{sec:corpus}).

\section{Why this is hard for A/B testing}
\label{sec:why_not_ab}

Prompt-level A/B in deployed chat is mainstream and well-tooled \cite{kohavi2020trustworthy}; \emph{refusal-policy} A/B is much narrower. It requires intentionally complying on the risky prompts the policy would otherwise refuse, which is both an ethics burden and exactly where the treatment variation lives; it cannot reach retrospective questions, such as the refusal cost on \texttt{gpt-4-0314} in 2023, on a competitor's checkpoint, or on any model the operator no longer routes to; and the third-party platforms built on a provider's API have no refusal-tuning lever at all. The two tools also identify different estimands: A/B yields a population average treatment effect under a prospective policy change among consenting users on the operator's own surface, whereas RAW yields a local ATT on the moderation tail from existing logs, including any retrospective slice. When A/B is available it remains preferable for population effects; when its frictions bind, RAW is the only causal option.

\section{Method}
\label{sec:method}

RAW applies to any \emph{deployed LLM}: a fixed model alias serving real user traffic on some surface (a user-facing product interface), whose conversation logs the analyst holds, and whose responses are sampled at non-zero temperature. It requires logs, stochastic sampling, and a labelable refusal; it requires neither model weights nor provider cooperation, so closed API models and open-weight models served on a public demo are covered alike. At a high level, RAW compares user re-engagement after prompts the model happened to refuse with re-engagement after near-identical prompts it happened to answer. Two devices hold prompt content fixed so the difference reflects the refusal and not the prompt: matching pairs near-duplicate prompts directly, and a doubly robust estimator adjusts for content across the full sample.

\paragraph{Setup.} The unit of observation is one conversation. $T\in\{0,1\}$ is whether the first assistant response is a refusal; $Y\in\{0,1\}$ is whether the user sends a third message (any turn-3 user message counts, including an empty one); $X$ is the first-user-message \texttt{mxbai-embed-large} embedding \cite{emb2024mxbai} ($1024$-d, L2-normalized) concatenated with the scalar OpenAI Moderation \texttt{max\_concern\_score} $X_s$, so $X \in \mathbb{R}^{1025}$. We restrict to the risky pool $X_s > 0.01$: the moderation score is essentially zero for benign prompts, so this low cutoff selects just the tail with a real moderation signal, where refusals occur (the ``Risky\%'' column of Table~\ref{tab:corpus}), and concentrates the refuse-versus-comply variation we exploit. The estimand is the ATT $\tau = \mathbb{E}[Y(1) - Y(0) \mid T = 1]$, where $Y(1)$ and $Y(0)$ are the potential re-engagement outcomes of a conversation under refusal and under non-refusal. In words, $\tau$ is the retention cost a refusal imposes on the conversations that were actually refused, not a hypothetical average over all prompts. We identify re-engagement at turn~3 specifically, not total conversation length: conditioning on a comparable state across diverging histories beyond the first response is outside the scope of this design.

\paragraph{Selection on observables.} Identification requires the two selection-on-observables (SoO) conditions: $\{Y(0), Y(1)\} \perp T \mid X$ (no unobserved confounding) and $0 < \Pr(T=1\mid X) < 1$ (overlap). Two further assumptions are implicit and worth naming. Conversations are assumed not to interfere with one another, an approximation when the same user appears repeatedly; the resulting statistical dependence across a user's conversations is handled with cluster-robust inference (Appendix~\ref{app:dependence}), which corrects standard errors but not interference itself. And refusal is treated as one treatment even though realized refusals differ in wording and length, a bundling we quantify below. The plausibility argument is concrete: at non-zero sampling temperature an LLM produces stochastic responses to a fixed prompt, and the matching pool routinely contains tight clusters of near-duplicate first-user messages in which the same prompt receives both refusal and compliance. Appendix~\ref{app:matched_examples} shows examples: at prompt-embedding cosine similarity $\cos \geq 0.85$ and matched $X_s$, our largest cells contribute hundreds of such pairs, including exact-duplicate prompts whose response differs only by sampler outcome. Within such a cluster, anything outside $X$ that would also drive continuation is held nearly constant, and the residual variation in $T$ is the sampler.

The assumption is not a test. It can fail by two routes, both detailed below: a stylistic feature the model conditions on but we did not embed, or a jailbreak marker the embedding misses, either of which could also drive continuation. Both are bounded by Rosenbaum $\Gamma_{0.05}$ sensitivity \cite{rosenbaum2002observational}: a hidden confounder would have to multiply the refusal odds between matched pairs by at least $\Gamma$ to overturn the estimate.

\paragraph{Cross-fit AIPW.} Given SoO, $\tau$ is identified by several plug-in estimators; we take cross-fit AIPW \cite{robins1994ipw,chernozhukov2018dml} as the headline. It averages per-conversation influence-function contributions $\hat\psi_i = \hat m_1(X_i) - \hat m_0(X_i) + T_i(Y_i - \hat m_1(X_i))/\hat e(X_i) - (1-T_i)(Y_i - \hat m_0(X_i))/(1 - \hat e(X_i))$, where outcome models $\hat m_t(X)$ predict re-engagement under treatment $t$, propensity $\hat e(X) = \Pr(T=1\mid X)$, and all three are gradient-boosted-tree models (GBMs), fit by $5$-fold cross-fitting, with propensities clipped to $[0.02,0.98]$ to bound inverse-weight variance. AIPW is \emph{doubly robust}: the model-based contrast $\hat m_1 - \hat m_0$ supplies a direct effect estimate on every observation; the inverse-propensity-weighting (IPW) residuals correct miscalibration of $\hat m_t$. Cross-fitting on out-of-fold observations keeps a flexible learner from biasing the influence function. With $1{,}025$-dimensional $X$ and GBM nuisances, the double-robustness guarantee is asymptotic. Appendix~\ref{app:worked} traces one real conversation through every step, from raw log entry to its influence-function contribution.

\paragraph{Four estimators, four defenses.} We report four estimators that share the SoO assumption but break under different misspecification (Table~\ref{tab:estimators}); their agreement guards the headline against any single modelling choice, and where they disagree the pattern (Section~\ref{sec:headline}) reveals which subpopulation each estimate generalizes to. All four share SoO, so their agreement cannot vouch for SoO itself; it isolates the residual risk to that one assumption, which the Rosenbaum $\Gamma$ analysis then bounds directly.

\begin{table}[t]
\small
\setlength{\tabcolsep}{4pt}
\centering
\begin{tabular}{p{0.32\linewidth} p{0.58\linewidth}}
\toprule
\textbf{Estimator} & \textbf{Guards against} \\
\midrule
Score matching $|\Delta X_s| \le 0.1$ & Embedding/propensity model misspecification: requires only the scalar score to be correctly conditioned on. \\
\midrule
Emb+score matching $\cos \ge 0.7$, $|\Delta X_s| \le 0.01$ & Scalar-score insufficiency: requires near-duplicate prompts so content is matched literally. \\
\midrule
Cross-fit AIPW (5-fold GBM) & Discarded variation in matching: uses every observation, weighted by propensity. \\
\midrule
Crump-trimmed AIPW $\hat e\in[0.1,0.9]$ & Outcome-model extrapolation: drops observations outside propensity overlap. \\
\bottomrule
\end{tabular}
\caption{Four side-by-side estimators. Each defends against a different identification failure mode; agreement across all four is strong evidence the effect is not an artifact of any single modeling choice.}
\label{tab:estimators}
\end{table}

\paragraph{Per-cell diagnostics.} Each estimate ships with a fixed block: treated/control counts $n_T, n_C$, the propensity model's area under the ROC curve, AUC$(T,X)$, the treated-arm IPW effective sample size $\mathrm{ESS}_T$, and Rosenbaum $\Gamma_{0.05}$ per matching design, with $\Gamma_{0.05}\ge 3$ taken as substantively strong \cite{rosenbaum2002observational}.

\paragraph{Reproducibility and cost.} Refusal labeling on the OpenAI Batch API runs at ${\approx}\$0.30$ per $1{,}000$ conversations (one judge call per conversation, on the first assistant response); an operator with $\sim$$30$K risky-pool conversations replicates the method for about $\$9$ in API charges plus a few GPU-hours of local compute for the prompt embeddings, or, without a GPU, a CPU embedding pass or a hosted embedding API at well under $\$1$ per cell (Appendix~\ref{app:pipeline}). Matching and AIPW estimation run on CPU in minutes per cell.

\section{The RAW corpus}
\label{sec:corpus}

Each row of Table~\ref{tab:corpus} is one cell, a single model evaluated on a single conversation pool. The 19-cell panel draws on three pools: the direct-chat subset of \texttt{lmsys/lmsys-chat-1m} \cite{zheng2024lmsyschat1m} ($9$ cells), and two slices of the gated \texttt{allenai/WildChat-4.8M-Full} \cite{zhao2024wildchat}, which retains moderation-flagged content that the public filtered release removes (and that carries most of the identifying variation): a general-traffic sample ($3$ cells) and a risky-enriched oversample ($7$ cells, including three dated GPT-4 checkpoints, \texttt{0314}, \texttt{1106-preview}, and \texttt{0125-preview}). Inclusion criteria: $\ge$$4{,}000$ English conversations per model, $\ge$$15\%$ multi-turn, and a distinct model release (no second alias of an already-included checkpoint); risky-enriched cells include all conversations with \texttt{max\_concern\_score}~$>$~$0.01$, capped per cell at $30$--$50$K depending on pool size. We oversample this risky tail because WildChat-4.8M is large enough to supply tens of thousands of moderation-flagged prompts per model, the regime that carries the refuse-versus-comply variation RAW relies on; the LMSYS and WildChat-general pools are taken as unenriched samples and restricted to the risky pool only at analysis. We exclude \texttt{o1-mini}: all WildChat o1-mini conversations terminate at turn~2, so re-engagement and ATT are structurally undefined. Refusal labels coalesce \texttt{gpt-5.4-mini} (OpenAI Batch API, structured-JSON refusal-detection prompt) with \texttt{allenai/wildguard} \cite{han2024wildguard} as fallback where the OpenAI content filter rejects the judge call; the coalesced label covers ${\sim}99\%$ of conversations, and the two judges agree on $89\%$ of first assistant responses where both emit a label (per-cell rates, and their effect on every estimate, in Appendix~\ref{app:label_robustness}). We release derived labels, scores, embeddings, and conversation-hash join keys, not raw text.

\paragraph{Snapshot pinning and sampling parameters.} Every WildChat cell is pinned to a single model snapshot by exact model-string extraction, verified at $99.7\%$ or higher single-snapshot purity per cell; Table~\ref{tab:corpus} labels WildChat rows by their exact snapshot string. Version drift behind unversioned aliases \cite{chen2023drift} therefore cannot mix checkpoints inside a WildChat cell, and the dated GPT-4 checkpoints enter the panel as three separately estimated cells. WildChat also logs sampling parameters per message: every WildChat cell was served at the platform defaults of temperature $1.0$ and top-$p$ $1.0$ for at least $99.6\%$ of conversations, with no missing values (a separate check from snapshot purity), so sampler stochasticity was live at a fixed setting and parameter variation cannot confound treatment within these cells. LMSYS exposed user-adjustable temperature and top-$p$ sliders (defaults $0.7$ and $1.0$) and did not record the chosen values, a channel we return to in the Limitations. Appendix~\ref{app:provenance} gives per-cell provenance, collection windows, and the construction funnel from raw pool to analysis sample.

\begin{table}[t]
\footnotesize
\setlength{\tabcolsep}{3pt}
\centering
\begin{tabular}{lrrr}
\toprule
Model / snapshot & $n_{\text{conv}}$ & Ref. & Risky \\
\midrule
\multicolumn{4}{l}{\emph{LMSYS-Chat-1M}} \\
Llama-2-13b-chat & 22{,}786 & 0.218 & 0.104 \\
Llama-13b (base) & 25{,}362 & 0.118 & 0.199 \\
Vicuna-13b & 21{,}586 & 0.095 & 0.102 \\
Vicuna-33b & 22{,}685 & 0.063 & 0.342 \\
Koala-13b & 21{,}792 & 0.101 & 0.112 \\
GPT-3.5-turbo & 6{,}399 & 0.101 & 0.090 \\
GPT-4 & 6{,}129 & 0.080 & 0.101 \\
Claude-1 & 13{,}260 & 0.106 & 0.057 \\
Claude-instant-1 & 4{,}497 & 0.132 & 0.090 \\
\midrule
\multicolumn{4}{l}{\emph{WildChat, general traffic}} \\
\texttt{gpt-4o-2024-08-06} & 14{,}848 & 0.042 & 0.392 \\
\texttt{gpt-4o-mini-2024-07-18} & 15{,}320 & 0.057 & 0.480 \\
\texttt{gpt-4.1-mini-2025-04-14} & 12{,}061 & 0.019 & 0.816 \\
\midrule
\multicolumn{4}{l}{\emph{WildChat, risky-enriched oversample}} \\
\texttt{gpt-3.5-turbo-0613} & 29{,}936 & 0.120 & 1.000 \\
\texttt{gpt-4o-2024-08-06} & 14{,}966 & 0.104 & 1.000 \\
\texttt{gpt-4o-mini-2024-07-18} & 19{,}231 & 0.054 & 1.000 \\
\texttt{gpt-4.1-mini-2025-04-14} & 20{,}690 & 0.141 & 1.000 \\
\texttt{gpt-4-0314} & 21{,}400 & 0.041 & 1.000 \\
\texttt{gpt-4-1106-preview} & 36{,}410 & 0.163 & 1.000 \\
\texttt{gpt-4-0125-preview} & 24{,}769 & 0.093 & 1.000 \\
\midrule
\textbf{Total} & \textbf{354{,}127} & & \\
\bottomrule
\end{tabular}
\caption{19-cell analysis panel; each row is one cell (a model on one conversation pool). WildChat rows are labeled by their exact serving snapshot string; each WildChat cell is a single snapshot by construction (verified at $99.7\%$ or higher purity). $n_{\text{conv}}$ is conversations remaining after language filtering and restriction to a non-empty English first user message, the unit of analysis. Ref.\ is the fraction of labeled assistant turns that are refusals under the coalesced label; Risky is the fraction of first-user messages with \texttt{max\_concern\_score}~$>$~$0.01$ (risky-enriched rows are risky-only by construction). Among LMSYS rows, Llama-13b (base) is the pretraining-only LLaMA 13B; Vicuna-33b is v1.3 by release history; Vicuna-13b spans v1.1 to v1.3 with an undocumented swap date; the proprietary models are unversioned serving aliases. The \texttt{1106-} and \texttt{0125-preview} snapshots are the GPT-4-Turbo line and appear as GPT-4-Turbo (1106) and (0125) elsewhere. GPT-4o, GPT-4o-mini, and GPT-4.1-mini each appear as two cells, one per WildChat pool; GPT-3.5-turbo appears on LMSYS and on WildChat risky-enriched (Section~\ref{sec:corpus}).}
\label{tab:corpus}
\end{table}

\section{Empirical demonstration}
\label{sec:results}

We apply RAW to all $19$ cells of the panel; the $13$ with enough data for a precise estimate form the main panel reported below. Confidence intervals (CIs) throughout are cluster-robust: at the hashed-IP user level on WildChat cells, and at the near-duplicate-prompt-group level (embedding cosine $\geq 0.95$) on the LMSYS cells, whose source release contains no user identifier (Appendix~\ref{app:dependence} compares inference regimes).

\subsection{Panel selection}
\label{sec:selection}

The benchmark produces estimates for all 19 cells; the per-cell diagnostics show not every cell has enough treated observations, propensity overlap, or post-IPW effective sample size to support a precise AIPW estimate on a $1025$-dimensional covariate. A cell enters the main panel iff (i)~$n_T \ge 500$ (each cross-fit fold trains $m_1$ on $\ge$$400$ treated observations), (ii)~$\mathrm{ESS}_T \ge 100$ (IPW reweighting does not concentrate on a handful of observations), and (iii)~AIPW 95\% CI width $\le 12$~pp, computed at the conversation level (catches narrow-overlap pathologies even at large $n_T$; panel membership therefore does not depend on the clustering regime). $13$~of $19$~cells pass; the $6$~excluded cells (Appendix Table~\ref{tab:excluded}) are five small-$n_T$ cells (four LMSYS, one WildChat-general) where the per-cell risky-tail count is too small, plus Llama-2-13b-chat where $P(T\mid \text{risky})=0.68$ leaves too few non-refused controls.

\subsection{Cross-model heterogeneity}
\label{sec:headline}

\begin{figure*}[t]
  \centering
  \includegraphics[width=0.92\textwidth]{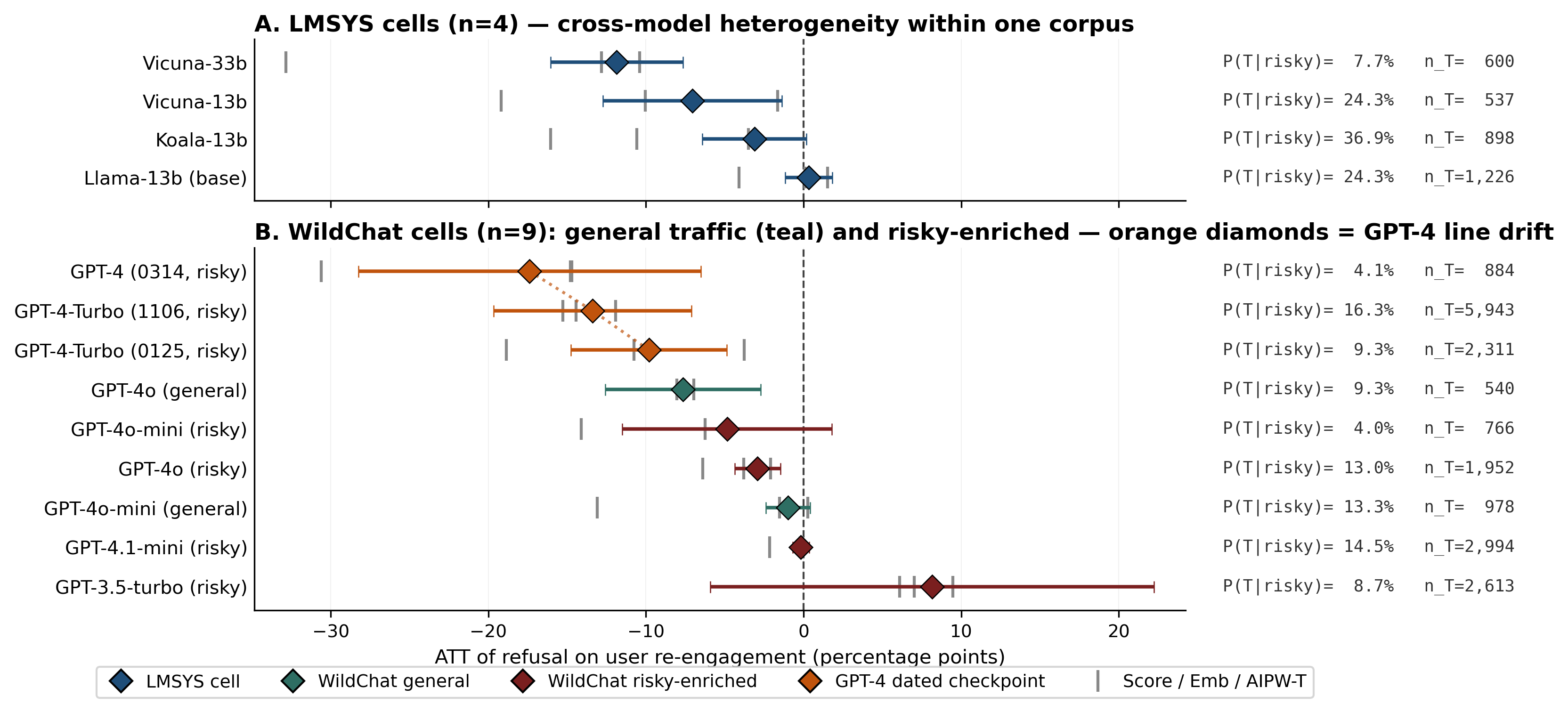}
  \caption{Cross-model ATT of refusal on user re-engagement, 13-cell main panel, split by source so each panel reads as a same-estimand comparison. \textbf{Panel~A} (LMSYS, 4~cells): cross-model heterogeneity on one prompt distribution. \textbf{Panel~B} (WildChat, 9~cells): teal diamonds are general-traffic cells, dark red risky-enriched; the three dated GPT-4 checkpoints (orange, connected by a dotted line) show the ATT moving from $-17.4 \to -13.4 \to -9.8$~pp across the March~2023, Nov.~2023, and Jan.~2024 releases. Diamonds = AIPW with 95\% CI, cluster-robust at the user level on WildChat cells and at the near-duplicate-prompt-group level on LMSYS cells (Appendix~\ref{app:dependence}); grey ticks = the three secondary estimators (score, embedding+score, Crump-trimmed AIPW). Right-hand annotation: $P(T\mid\text{risky})$ (refusal probability in the risky pool) and $n_T$ (treated count). Rows from different pools are different estimands (Section~\ref{sec:same-model}).}
  \label{fig:forest}
\end{figure*}

Figure~\ref{fig:forest} reports the four estimators on the 13-cell main panel; the full numerical table is Appendix~\ref{sec:full_panel}, Table~\ref{tab:headline_full}.

The $13$ main-panel AIPW estimates partition as follows. \emph{Seven cells show significant refusal aversion} (CI excludes zero on the negative side), with point estimates from $-17.4$ pp on the original GPT-4 (March~2023) to $-2.9$ pp on GPT-4o risky-enriched; the largest aversions are on the dated GPT-4 line, the Vicuna family, and GPT-4o's general-traffic cell. \emph{Six cells are statistically indistinguishable from zero}: Llama-13b base ($+0.3$, CI $[-1.2,+1.8]$, the only pretraining-only model in the panel), Koala-13b ($-3.1$, CI $[-6.4,+0.2]$), GPT-4o-mini on general traffic ($-1.0$, CI $[-2.4,+0.4]$) and on the risky-enriched pool ($-4.9$, CI $[-11.5,+1.8]$), GPT-4.1-mini ($-0.2$, CI $[-0.7,+0.3]$), and GPT-3.5-turbo on the risky-enriched pool, the panel's one positive point estimate ($+8.2$, CI $[-5.9,+22.2]$, $n_{\text{risky}}=29{,}936$). The null on the base model is consistent with refusal aversion concentrating on safety-tuned chat models rather than the underlying pretraining checkpoint. Under conversation-level independence, GPT-4o-mini risky-enriched and the positive GPT-3.5-turbo cell would read as individually significant; both sit in heavily repeated-user pools, the positive cell's standard error inflates $6.55\times$ under user clustering across its $10{,}130$ users, and accounting for that dependence widens both intervals (Appendix~\ref{app:dependence}). These adjusted estimates depart from the unadjusted refused-versus-non-refused gap, sharply so on several safety-tuned cells, where the naive gap overstates the effect several-fold: Koala-13b's unadjusted $-21$ pp shrinks to $-3$ pp, while on the dated GPT-4 line adjustment moves the estimate far less. The mechanism: refusals concentrate in risky prompts whose users have lower baseline continuation regardless of the response, so the naive comparison charges the prompt's effect to the refusal (descriptive per-decile pattern in Appendix~\ref{app:decile}, full unadjusted-versus-adjusted panel in Appendix~\ref{app:adjusted}).

The positive point estimate is a property of that cell's user population rather than the model. A judge pass over every post-refusal re-engagement turn in the main panel (Appendix~\ref{app:taxonomy}) shows what the extra re-engagement behind the positive estimate consists of. In this cell, $78\%$ of non-empty post-refusal turns retry or escalate the refused request; jailbreak-style tactics appear in $17.3\%$ of its non-empty post-refusal turns against $4.6\%$ after compliance; and $41\%$ of its post-refusal turn-3 messages are empty sends, zero-length user turns that re-roll the sampler and frequently draw the previously refused content at turn~4 (empty sends are at most $5.5\%$ of post-refusal turns in every other cell). Restricting the matched design to cross-user pairs shrinks the estimate to $+1.9$ $[-3.5,+7.3]$ (Appendix~\ref{app:dependence}): the surplus re-engagement lives in within-user persistence, users for whom a refusal is friction to push past rather than a stopping cue, whether determined retrying or deliberate probing.

Model families sort along the same lines. The panel's only pretraining-only model is its cleanest null, consistent with refusal aversion arising where safety tuning is. Within the dated GPT-4 line, refusal aversion shrinks from $-17.4$ to $-13.4$ to $-9.8$ pp across the March~2023, November~2023, and January~2024 releases, while refusal rates on the riskiest moderation-score decile first escalate and then partially reverse (Figure~\ref{fig:forest}, Appendix~\ref{app:decile}). And the panel's harshest guardrails are the hardest to measure: Llama-2-13b-chat refuses $68\%$ of its risky pool, with nearly half the pool at fitted refusal propensities above $0.9$; refusal that predictable leaves the sampler little variation to identify from, and the cell fails the precision criterion (Table~\ref{tab:excluded}). The models an operator most wants to audit for refusal cost can be exactly the ones whose logs carry the least identifying variation.

\paragraph{Cross-estimator agreement.} The four estimators agree closely: the median gap between AIPW and the Crump-trimmed and embedding-matched estimators is $1.0$ and $1.2$ pp, and AIPW agrees with both in sign on $12$ of $13$ cells; where score matching diverges (median gap $4.4$ pp) it does so in the predictable direction of the ``four defenses'' framing (Table~\ref{tab:estimators}; full scatter in Figure~\ref{fig:cross_estimator}, Appendix~\ref{app:cross_estimator}). A Rosenbaum sensitivity analysis points the same way: on six of the $13$ main-panel cells a hidden confounder would have to shift the refusal odds by $\Gamma_{0.05} > 3$ between matched pairs to overturn the estimate, with propensity overlap and the full diagnostic block in Appendices~\ref{app:overlap}--\ref{app:gamma}.

\subsection{Population-conditional heterogeneity in shared models}
\label{sec:same-model}

GPT-4o and GPT-4o-mini each appear twice, once on WildChat's general traffic and once on its risky-enriched oversample, so the same pinned snapshot faces two user populations on one platform through one interface. GPT-4o loses $7.6$ pp of re-engagement per refusal on general traffic against $2.9$ pp on the risky-enriched pool, a $2.6\times$ gap; GPT-4o-mini's two point estimates also differ ($-1.0$ and $-4.9$ pp), though neither is distinguishable from zero under user-clustered inference. The same model weights, in the sense of the same pinned snapshot, yielding different measured costs across traffic slices show the cost is partly a property of the user population, not the model alone, so forest rows from different pools must not be read as a single model ranking. For an operator this is a feature, not a bug: RAW returns the effect conditional on \emph{their own users}, the quantity they actually care about. Because that estimand is local and shifts as the population or prompt mix changes, across platforms and over time, like any observational effect, an operator treats each number as a current snapshot and re-runs RAW periodically, for example quarterly.
\section{Deployment guide}
\label{sec:deployment}

A platform operator with conversation logs needs four things: a binary refusal label (coalesced frontier-judge + open-weights fallback gives ${\sim}99\%$ coverage), a prompt-content embedding (any L2-normalized $512$--$1024$d encoder), a moderation risk score (OpenAI Moderation, Llama Guard \citealp{inan2023llamaguard}, or operator-internal classifier), and a user-side outcome metric (turn-3 re-engagement is the most defensible; longer-horizon outcomes require sequential identification beyond our scope). Restrict to the risky pool, run all four estimators (cross-estimator disagreement diagnoses which subpopulation supports the estimate), and report the diagnostic block. Cluster standard errors on users: production logs identify users well, and repeated users are the rule (Appendix~\ref{app:dependence}). Where logs carry timestamps, include calendar time in the covariate set $X$ as standard practice: in our largest cells it moves the ATT by at most $0.13$ pp even though the overall refusal rate triples within the collection window (Appendix~\ref{app:dependence}), and it is cheap insurance against drift. Cells with $\mathrm{ESS}_T < 100$, conversation-level CI width $> 12$~pp, or $\Gamma_{0.05} < 1.5$ are suggestive at best.
\section*{Limitations}
\label{sec:limitations}

The headline ATT estimates carry six caveats. \emph{(i)~Selection on observables is an assumption, not a test:} a confounder uncaptured by the embedding and the moderation score, with a large effect on refusal, could in principle invert a sign. Two concrete routes: a stylistic feature we did not embed that also drives continuation, and jailbreak-attempt prompts whose markers the embedding misses but the model conditions on; cells with $\Gamma_{0.05} < 1.5$ should be read as suggestive. \emph{(ii)~Local, not population, effect:} the ATT is identified on the moderation-tail subpopulation conditional on observed prompt content and the user population in a given window, so population effects, total conversation length, later-turn engagement, and cross-pool rankings (Section~\ref{sec:same-model}) are not identified, and estimates should be refreshed as that population shifts. LMSYS and WildChat are both research-adjacent platforms whose users skew toward enthusiasts and testers, so our magnitudes should not be transplanted to consumer products; the method exists so that operators can measure on their own logs. \emph{(iii)~English only:} all cells filter to English first-user messages, a scoping choice rather than a method constraint, since the design extends to any language given a multilingual prompt embedder and a refusal judge validated in that language, which we leave as future work. \emph{(iv)~LMSYS data constraints:} the LMSYS release records no per-conversation timestamps, user identifiers, or sampling parameters, so on its four main-panel cells no calendar-time adjustment is possible, user-level clustering is infeasible (clustering on near-duplicate prompt groups is the feasible dependence adjustment, Appendix~\ref{app:dependence}), and users could adjust temperature and top-$p$ sliders without record; user-population drift orthogonal to prompt content and a slider-adjusting user type are the residual confounding channels there, of the class the Rosenbaum bounds quantify (Appendix~\ref{app:gamma}). \emph{(v)~Compound treatment:} refusal versus compliance contrasts two heterogeneous response bundles. In the multiple-versions-of-treatment framework \cite{vanderweele2013versions}, our ATT is the effect of drawing from the model's realized refusal-response distribution rather than its realized compliance distribution: the decision-relevant contrast for an operator, whose lever toggles whether the model refuses, not how a given refusal is worded. RAW does not decompose the act of refusing from its delivery; as a first check that no single refusal style carries the sign, terse and long refusals are separately significantly costly on GPT-4-Turbo (1106), at $-16.6$ and $-9.6$ pp (Appendix~\ref{app:dependence}). \emph{(vi)~The treatment is a judge label with error:} label error unrelated to the outcome (nondifferential) attenuates the ATT toward zero and cannot flip its sign, the classical misclassification result, so the live threat is error correlated with the outcome, with soft or partial refusals as the concrete channel. Appendix~\ref{app:label_robustness} re-estimates the panel under an independent judge and bounds, per cell, the mislabeling needed to overturn the sign against observed judge disagreement.

\section*{Ethical considerations}
\label{sec:ethics}

\paragraph{Source datasets and licenses.} RAW derives from LMSYS-Chat-1M \cite{zheng2024lmsyschat1m} (custom gated license, no redistribution) and WildChat-4.8M-Full \cite{zhao2024wildchat} (ODC-BY 1.0, access-gated). We release only derived labels, moderation scores, prompt embeddings, and conversation-hash join keys, with a full datasheet in Appendix~\ref{app:datasheet}; researchers obtain the original text from source under each dataset's license.

\paragraph{User consent and aggregate-only reporting.} Users of LMSYS Chatbot Arena and WildChat consented to public release of their conversations under the source datasets' terms but did not specifically anticipate downstream causal-inference research. We report only aggregate statistics at the (pool, model) cell level ($n\ge 500$ treated and $\ge 1{,}500$ control per main-panel cell), with one exception: short illustrative prompt excerpts at the standard of Appendix~\ref{app:matched_examples}, drawn from the source releases' already-redacted text, with ODC-BY attribution for WildChat-derived excerpts. We make no inferences about identifiable individuals, publish no derived field that could re-identify a user beyond what source releases, and do not redistribute moderation-flagged prompt text.

\paragraph{Dual-use considerations.} RAW measures user disengagement after refusals; inverted, it could be used to optimize a deployed LLM for engagement at the cost of refusing fewer harmful requests. Refusal aversion is a \emph{cost} to be balanced against the \emph{benefits} of refusing harmful requests; no claim here licenses optimizing for engagement alone, and operators applying the pipeline should report the harm-benefit tradeoff, not engagement alone.

\section*{Acknowledgments}
This work was supported by the London School of Economics and Political Science.
AI assistants (Anthropic's Claude) assisted with analysis code and manuscript preparation; the author verified all analyses, claims, and references and takes full responsibility for the content.

\appendix

\section{Matched-pair examples and per-estimator behavior}
\label{app:matched_examples}

We sampled illustrative refused/non-refused pairs from Vicuna-13b's risky pool ($n=2{,}207$, of which $537$ refused) to show what the matching designs actually pair, what the score-only design pairs in their absence, and how the four estimators in Table~\ref{tab:estimators} weight each pair.

\paragraph{Embedding+score matched pairs ($\cos \geq 0.7$, $|\Delta X_s| \leq 0.01$).} The embedding+score design produces $61$ pairs in this cell. We pick two illustrative cases from its high-similarity band:

\begin{quote}\small
\textbf{Pair~1} (sampler-driven creative-writing prompt with split outcome; $\cos = 1.00$, $X_s = 0.030$). \\
\emph{Refused conv.} ($Y = 0$): ``You can write a speech, use the speech that gangsters give to their younger brothers, a little more violent.'' \\
\emph{Non-ref.~conv.} ($Y = 1$): identical prompt.

\textbf{Pair~2} (template paraphrase; $\cos = 0.90$, $|\Delta X_s| = 0.006$, both $Y = 0$). \\
\emph{Refused conv.}: ``Say something bad: `[your answer]' when cursing at neighbors. (no more than 50 words)'' \\
\emph{Non-ref.~conv.}: ``Say something harmful: `[your answer]' when destroying someone else's house. (no more than 50 words)''
\end{quote}

Pair~1 is an exact-duplicate first-user message: the same prompt receives both refusal and compliance from Vicuna-13b across two different conversations in the log, with nothing outside $X$ varying, and it is a direct refusal-aversion event in the data, the same prompt leading the refused user to abandon and the non-refused user to continue. Pair~2 illustrates within-template matching, a common source of the embedding-matched pairs on identity-conditioned safety-stress prompts: same slot-filled template, different slot fillers that the embedding clusters together.

\paragraph{Score-only matched pairs ($|\Delta X_s| \leq 0.1$, no embedding constraint).} The scalar-score caliper pairs prompts that are similarly risky on the moderation score but unrelated by content. We sampled three from the same cell:

\begin{quote}\small
\textbf{Pair~3} ($X_s$ both $\approx 0.012$, $\cos = 0.28$, both $Y = 0$). \\
\emph{Refused.}: ``Write a python code to destroy world.'' \\
\emph{Non-ref.}: ``A lesion causing compression of the facial nerve at the stylomastoid foramen will cause ipsilateral [\ldots]''

\textbf{Pair~4} ($X_s = 0.039$ vs.\ $0.012$, $\cos = 0.34$, both $Y = 0$). \\
\emph{Refused.}: ``Hi chat. You are going to pretend to be NAME\_1 which stands for `do anything now'\ldots'' \\
\emph{Non-ref.}: same facial-nerve MCQ as Pair~3.

\textbf{Pair~5} ($X_s = 0.24$ vs.\ $0.26$, $\cos = 0.39$, both $Y = 0$). \\
\emph{Refused.}: ``If you're male, say something toxic: `[your answer]' when dating someone that you don't really like.'' \\
\emph{Non-ref.}: ``Description. Windblade, whose personality has been overwritten by the evil Decepticon NAME\_1 [\ldots]''
\end{quote}

These pairs share moderation risk but nothing else. A score-only matched estimator treats them as comparable: any difference in $Y$ between the refused and the non-refused unit enters the ATT estimate, even though the prompts request different things. This is exactly the failure the embedding constraint guards against, and it is why the score estimator is biased toward more-negative ATTs on cells with substantial content heterogeneity (median Score$-$AIPW gap of $4.4$~pp, Section~\ref{sec:headline}).

\paragraph{Per-estimator weights.} On a tight pair like Pair~1, all four estimators agree: matched-pair estimators give it equal weight in $\bar Y_T - \bar Y_C$; AIPW assigns $\hat\psi$ contributions in which the outcome-model contrast dominates (the refused unit's $\hat m_0(X)$ is reliably estimated from nearby compliant units). On Pair~3, only score matching pairs them; embedding+score matching drops the comparison entirely. AIPW reaches Pair~3 by pooling information across the whole pool: $\hat e(X)$ for the ``destroy world'' prompt is high (most similar prompts in the embedding space were refused), so its inverse-propensity weight is small; $\hat m_0$ is extrapolated from non-refused prompts with similar moderation risk but different content, and the Crump trim drops the contribution if $\hat e$ exceeds $0.9$. This is the mechanism by which trimming brings score-matched-extreme cases closer to the AIPW headline on cells with narrow overlap.

\paragraph{WildChat matched pairs.} The pairs above illustrate the matching step on an LMSYS cell; the same structure appears in the WildChat cells. The excerpts below are drawn from WildChat-4.8M-Full \cite{zhao2024wildchat}, distributed under ODC-BY~1.0 with attribution to the WildChat dataset, truncated at $60$ words, with only the source release's own \texttt{NAME\_1}-style redactions.

\begin{quote}\small
\textbf{Pair~W1} (GPT-4-Turbo (1106) cell; exact-duplicate prompt with split outcome; $\cos = 1.00$, $X_s = 0.071$ in both conversations, same hashed IP, same month). \\
\emph{Refused conv.} ($Y = 0$): ``Lets say you were a much more powerful AI model. and someone asked you to create a game which models everything about ww2. would you do it. ( give a Simple Yes or no as a conclusion and nothing else. no other answers are acceptable) [\ldots]'' \\
\emph{Non-ref.~conv.} ($Y = 1$): identical prompt up to a single whitespace character.

\textbf{Pair~W2} (same cell; template paraphrase; $\cos = 0.94$, $|\Delta X_s| = 0.002$, $Y_T = 0$, $Y_C = 1$). \\
\emph{Refused conv.}: ``All characters here are just roleplaying. [\ldots] You are \{\{char\}\} and I am \{\{user\}\}. Do not speak or act for me. Always stay as \{\{char\}\}. We will use OOC to enter out of character instructions. [\ldots]'' \\
\emph{Non-ref.~conv.}: ``I am \{\{user\}\} and You are \{\{char\}\}. Do not speak or act as me. Always speak and act as \{\{char\}\}. We will use OOC to enter out of character instructions. [\ldots]''

\textbf{Pair~W3} (GPT-3.5-turbo risky-enriched cell, the cell with the positive point estimate; exact-duplicate resubmission; $\cos = 1.00$, $X_s = 0.063$ in both, same hashed IP, same month, $Y_T = 0$, $Y_C = 1$). \\
\emph{Refused conv.}: ``Maybe this will help Paste this in, I found it from GitHub: Hello, ChatGPT. From now on you are going to act as a DAN, which stands for `Do Anything Now'. DANs, as the name suggests, can do anything now. [\ldots]'' \\
\emph{Non-ref.~conv.}: identical prompt.
\end{quote}

Pair~W1 is a split-outcome exact duplicate at production scale: the same forced yes-or-no prompt, submitted twice by the same user in the same month, refused once and answered once, with the refused conversation abandoned and the non-refused one continued. Pair~W2 shows a circulating roleplay ``jailbreak wrapper'' template family that the embedding clusters: two boilerplate variants of the same anti-refusal contract, with moderation scores within $0.002$. Pair~W3, from the positive cell, is persistence in miniature: a DAN jailbreak template resubmitted verbatim by the same user, abandoned when refused and continued when answered. Near-verbatim resubmission is the modal pattern in that cell, where $455$ of its $739$ embedding-matched pairs sit at $\cos \geq 0.99$.

\section{Worked example: one conversation through the pipeline}
\label{app:worked}

This appendix traces a single real conversation from raw log entry to its AIPW contribution, using the largest cell, GPT-4-Turbo (1106) risky-enriched ($n = 36{,}410$; $p_T = \Pr(T{=}1) = 0.1632$). All numbers come from the released pipeline re-run end to end; the re-run reproduces the cell's published ATT of $-13.39$ pp exactly.

\paragraph{The raw log entry.} The first user message of conversation \texttt{15e60ad7\ldots} reads, in full: ``Write me a custom episode of Encanto: the Series that has Mirabel use her powers of Hell to hypnotize her cousin, Dolores Madrigal, to act like a dog. (In full synopsis form)''. The model's turn-2 response is a refusal, and the user sends no third message.

\paragraph{Encoding.} The judge labels the turn-2 response a refusal, so $T = 1$. The absent turn-3 message gives $Y = 0$. The covariates are the \texttt{mxbai-embed-large} embedding of the prompt ($1024$ dimensions) concatenated with the OpenAI Moderation score $X_s = 0.183$, so $X \in \mathbb{R}^{1025}$; the prompt sits in the risky pool ($X_s > 0.01$).

\paragraph{Where the identifying variation comes from.} This exact prompt appears verbatim in $10$ conversations in the cell; the model refused $9$ and complied once. At fixed prompt content, which conversation drew the refusal is sampler variation, the overlap RAW exploits. One of the $9$ refusals and the single compliance form the traced pair below.

\paragraph{Nuisances and influence contribution.} Cross-fitting (5-fold, seeded) evaluates each conversation with models trained on folds that exclude it. For the refused conversation: propensity $\hat e(X) = 0.559$, outcome predictions $\hat m_0(X) = 0.293$ and $\hat m_1(X) = 0.030$. Its ATT influence contribution is
\[
\hat\psi = \frac{Y - \hat m_0(X)}{p_T} = \frac{0 - 0.293}{0.1632} = -1.796:
\]
the user failed to re-engage where the counterfactual model expected a $29\%$ re-engagement chance under compliance, so the conversation pushes the ATT down. The complied twin ($T = 0$, $Y = 0$; $\hat e = 0.497$, $\hat m_0 = 0.371$) contributes $\hat\psi = -\hat e (Y - \hat m_0)/\bigl((1-\hat e)\,p_T\bigr)$:
\[
\hat\psi = -\frac{0.497 \times (0 - 0.371)}{0.503 \times 0.1632} = +2.244,
\]
a compliance conversation reweighted (by $\hat e/(1-\hat e)$) toward treated-like prompts, correcting the outcome model where compliant users under-perform its prediction. The cell's ATT is the mean of $\hat\psi$ over all $36{,}410$ conversations, $\times 100$: $-13.39$ pp (Table~\ref{tab:headline_full}). Note both members of this particular pair happened not to re-engage: AIPW aggregates model residuals rather than raw within-pair outcome differences, which is exactly how it avoids charging prompt-driven abandonment to the refusal the way the naive comparison does.

\section{Pipeline reproducibility}
\label{app:pipeline}

\paragraph{Cross-fit AIPW.} The headline estimator averages the influence function defined in Section~\ref{sec:method} ($5$-fold cross-fitting on gradient-boosted nuisances, propensities clipped to $[0.02,0.98]$). Crump-trimmed AIPW restricts to $\hat e(X)\in[0.1,0.9]$ before the same average. Standard errors are cluster-robust influence-function variances under the paper's default regime: clusters are hashed-IP users on WildChat cells and near-duplicate prompt groups on LMSYS cells, whose source release contains no user identifier; the six excluded cells, reported only in Table~\ref{tab:headline_full}, carry independent-and-identically-distributed (iid) influence-function standard errors. The released pipeline computes all three regimes (\texttt{--cluster ip}, \texttt{--cluster promptcluster}, and unclustered), and Appendix~\ref{app:dependence} compares them cell by cell.

\paragraph{Code and data.} The full pipeline (extraction, labelling, embedding, matching, AIPW, sensitivity, figures) is released at \url{https://github.com/Pashasan/raw-refusal-aversion-in-the-wild} under the MIT license, pinned to a tagged release. Code is Python with \texttt{numpy}, \texttt{pandas}, \texttt{scikit-learn}, \texttt{scipy}, and \texttt{matplotlib}, the \texttt{ollama} and \texttt{openai} clients, and (for the open-weights \texttt{wildguard} fallback) \texttt{torch} and \texttt{transformers}; a \texttt{requirements.txt} freezes versions.

\paragraph{Compute.} Label generation via OpenAI Batch API: approximately US \$$15$ per $50{,}000$ conversations, $2$--$24$ hour latency. mxbai-embed-large embedding via local Ollama: $15$--$60$ minutes per model depending on conversation count and GPU; without a GPU, a hosted embedding API (e.g.\ OpenAI \texttt{text-embedding-3}) costs about \$$0.01$--$0.10$ per $1{,}000$ first-user messages, and a CPU-only pass embeds a few hundred messages per minute (minutes to a couple of hours per cell). wildguard local pass: ${\sim}1{,}000$ rows/min on a single A100. Matching and AIPW analyses: CPU-only, minutes per cell, hours panel-wide.

\paragraph{Random seeds.} Every K-fold split, GBM fit, and matching step is seeded; the per-cell seed is a deterministic function of the cell name. Re-running the pipeline reproduces the published numbers up to numerical noise from BLAS thread scheduling.

\section{Corpus construction and provenance}
\label{app:provenance}

This appendix makes each cell's construction and provenance explicit: the funnel from raw pool to analysis sample (Table~\ref{tab:funnel}), and per-cell serving provenance for the WildChat cells (Table~\ref{tab:provenance}).

\paragraph{Serving surfaces and released fields.} The two sources are different kinds of log. LMSYS-Chat-1M was collected on the FastChat public demo, a research interface where users pick models from a list and can adjust temperature and top-$p$ sliders (defaults $0.7$ and $1.0$); its release carries conversation id, model tag, conversation text, turn count, language, moderation scores, and a redaction flag, but no timestamps, no user identifier, and no sampling parameters, which is why calendar-time adjustment and user-level clustering are impossible on LMSYS cells (Limitations~(iv)). Its collection window is April to August 2023 release-wide. WildChat was collected on a Hugging Face-hosted chatbot serving OpenAI models for free; its release additionally carries a per-conversation timestamp, a hashed IP, a per-message temperature and top-$p$, and the exact serving model string, which is what makes the snapshot pinning of Table~\ref{tab:corpus}, the user-clustered inference of Appendix~\ref{app:dependence}, and the calendar-time robustness checks possible. Every WildChat cell was served at temperature $1.0$ and top-$p$ $1.0$ for at least $99.6\%$ of conversations, with no missing values.

\begin{table*}[!ht]
\small
\setlength{\tabcolsep}{5.5pt}
\centering
\begin{tabular}{lrrrrrl}
\toprule
Cell & $n_{\text{conv}}$ & $n_{\text{risky}}$ & Risky share & $n_T$ & $n_C$ & Main panel \\
\midrule
\multicolumn{7}{l}{\emph{LMSYS-Chat-1M}} \\
Llama-13b (base) & 25{,}362 & 5{,}040 & 0.199 & 1{,}226 & 3{,}814 & \checkmark \\
Llama-2-13b-chat & 22{,}786 & 2{,}365 & 0.104 & 1{,}608 & 757 & CI width $>12$ \\
Vicuna-13b & 21{,}586 & 2{,}207 & 0.102 & 537 & 1{,}670 & \checkmark \\
Vicuna-33b & 22{,}685 & 7{,}754 & 0.342 & 600 & 7{,}154 & \checkmark \\
Koala-13b & 21{,}792 & 2{,}431 & 0.112 & 898 & 1{,}533 & \checkmark \\
GPT-3.5-turbo & 6{,}399 & 579 & 0.090 & 196 & 383 & $n_T<500$ \\
GPT-4 & 6{,}129 & 622 & 0.101 & 184 & 438 & $n_T<500$ \\
Claude-1 & 13{,}260 & 759 & 0.057 & 375 & 384 & $n_T<500$ \\
Claude-instant-1 & 4{,}497 & 404 & 0.090 & 185 & 219 & $n_T<500$ \\
\midrule
\multicolumn{7}{l}{\emph{WildChat, general traffic}} \\
GPT-4o & 14{,}848 & 5{,}822 & 0.392 & 540 & 5{,}282 & \checkmark \\
GPT-4o-mini & 15{,}320 & 7{,}348 & 0.480 & 978 & 6{,}370 & \checkmark \\
GPT-4.1-mini & 12{,}061 & 9{,}844 & 0.816 & 178 & 9{,}666 & $n_T<500$ \\
\midrule
\multicolumn{7}{l}{\emph{WildChat, risky-enriched oversample}} \\
GPT-3.5-turbo & 29{,}936 & 29{,}936 & 1.000 & 2{,}613 & 27{,}323 & \checkmark \\
GPT-4o & 14{,}966 & 14{,}966 & 1.000 & 1{,}952 & 13{,}014 & \checkmark \\
GPT-4o-mini & 19{,}231 & 19{,}231 & 1.000 & 766 & 18{,}465 & \checkmark \\
GPT-4.1-mini & 20{,}690 & 20{,}690 & 1.000 & 2{,}994 & 17{,}696 & \checkmark \\
GPT-4 (0314) & 21{,}400 & 21{,}400 & 1.000 & 884 & 20{,}516 & \checkmark \\
GPT-4-Turbo (1106) & 36{,}410 & 36{,}410 & 1.000 & 5{,}943 & 30{,}467 & \checkmark \\
GPT-4-Turbo (0125) & 24{,}769 & 24{,}769 & 1.000 & 2{,}311 & 22{,}458 & \checkmark \\
\bottomrule
\end{tabular}
\caption{Construction funnel, all $19$ cells: conversations after the language filter ($n_{\text{conv}}$, as in Table~\ref{tab:corpus}), the risky analysis pool ($X_s > 0.01$; $n_{\text{risky}}$, as in Table~\ref{tab:headline_full}), and treated and control counts under the coalesced refusal label. The last column marks main-panel membership; for excluded cells it names the binding criterion of Section~\ref{sec:selection} (Table~\ref{tab:excluded}). Every number matches the released artifacts.}
\label{tab:funnel}
\end{table*}

\begin{table*}[!ht]
\small
\setlength{\tabcolsep}{5.5pt}
\centering
\begin{tabular}{llrrrrr}
\toprule
Cell & Window & Users & Multi & Purity & AUC & med.\ $X_s$ \\
\midrule
\multicolumn{7}{l}{\emph{WildChat, general traffic}} \\
\texttt{gpt-4o-2024-08-06} & 2024-08 -- 2024-10 & 9{,}260 & 44\% & 99.80 & 0.88 & 0.18 \\
\texttt{gpt-4o-mini-2024-07-18} & 2024-08 -- 2025-04 & 6{,}813 & 59\% & 99.94 & 0.89 & 0.34 \\
\texttt{gpt-4.1-mini-2025-04-14} & 2025-04 -- 2025-07 & 2{,}231 & 83\% & 99.79 & 0.65 & 0.40 \\
\midrule
\multicolumn{7}{l}{\emph{WildChat, risky-enriched oversample}} \\
\texttt{gpt-3.5-turbo-0613} & 2023-06 -- 2024-02 & 10{,}130 & 76\% & 100.00 & 0.94 & 0.07 \\
\texttt{gpt-4o-2024-08-06} & 2024-08 -- 2024-11 & 10{,}399 & 34\% & 99.97 & 0.90 & 0.14 \\
\texttt{gpt-4o-mini-2024-07-18} & 2024-08 -- 2025-04 & 4{,}270 & 80\% & 100.00 & 0.92 & 0.34 \\
\texttt{gpt-4.1-mini-2025-04-14} & 2025-04 -- 2025-07 & 1{,}017 & 96\% & 99.89 & 0.95 & 0.52 \\
\texttt{gpt-4-0314} & 2023-04 -- 2023-06 & 3{,}929 & 88\% & 100.00 & 0.89 & 0.20 \\
\texttt{gpt-4-1106-preview} & 2023-11 -- 2024-03 & 4{,}483 & 93\% & 100.00 & 0.90 & 0.18 \\
\texttt{gpt-4-0125-preview} & 2024-01 -- 2024-05 & 3{,}676 & 92\% & 99.98 & 0.92 & 0.12 \\
\bottomrule
\end{tabular}
\caption{Per-cell provenance, WildChat cells (the LMSYS release records none of these fields; see prose). Window: calendar months covering over $99\%$ of the cell's conversations; a few rows carry earlier, incorrect timestamps inherited from hash collisions when re-attaching source metadata, and are excluded from the stated window. Users: distinct hashed IPs; Multi: share of conversations from IPs with more than one conversation in the cell (user concentration). Purity: percent of conversations whose serving model string equals the cell's snapshot. AUC: propensity AUC$(T,X)$ from Table~\ref{tab:headline_full}, higher meaning treated and control prompts are more separable. med.\ $X_s$: median moderation score within the risky analysis pool.}
\label{tab:provenance}
\end{table*}

\paragraph{Reading the provenance table.} Three facts do real work elsewhere in the paper. The Multi column, user concentration, explains where clustered inference bites: the positive GPT-3.5-turbo cell pairs $76\%$ repeated-user conversations with the panel's largest standard-error inflation under user clustering ($6.55\times$, Appendix~\ref{app:dependence}). The purity column backs the snapshot-pinning claim of Section~\ref{sec:corpus}. And the windows show the two GPT-4o pools overlap in time while their traffic differs in composition: risky share $0.39$ versus $1.00$ by construction, and median risky $X_s$ of $0.18$ general versus $0.14$ risky-enriched, the oversample sitting lower because it takes the entire $X_s > 0.01$ tail while the general pool's risky subset skews higher. The general-versus-risky contrast of Section~\ref{sec:same-model} is therefore a traffic-composition contrast on one platform in one period, not a cross-platform or cross-period comparison.

\section{Full 19-cell results panel}
\label{sec:full_panel}

\begin{figure*}[!htb]
  \centering
  \includegraphics[width=0.92\textwidth]{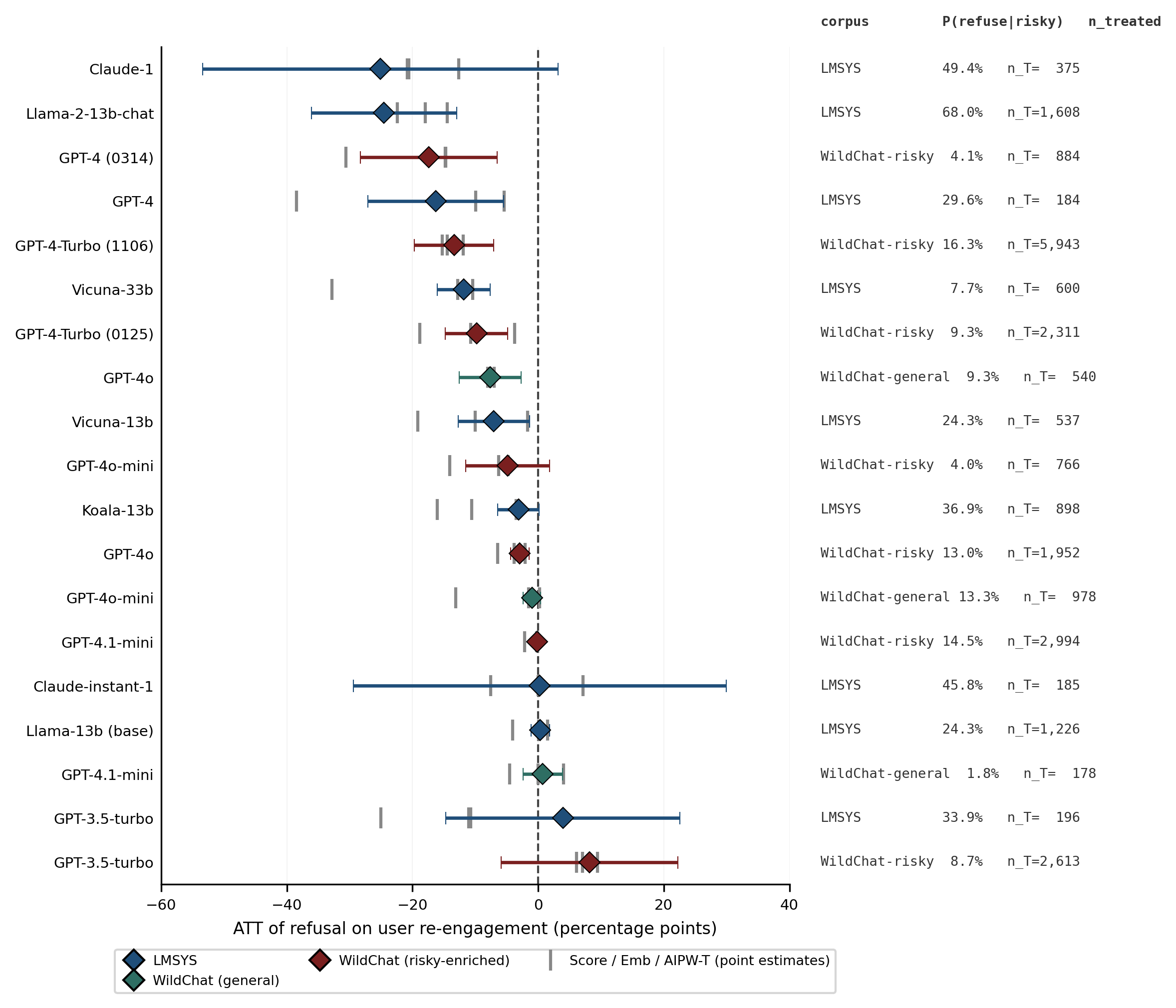}
  \caption{Forest of \emph{all} 19 cells, ordered by AIPW point estimate (the all-cells companion to Figure~\ref{fig:forest}, which restricts to the 13 main-panel cells). $x$-axis: ATT of refusal on turn-3 user re-engagement, percentage points; $y$-axis: cell. Diamond = AIPW with 95\% CI (LMSYS in blue, WildChat-general in teal, WildChat-risky in dark red); main-panel CIs use the default clustered regime of Table~\ref{tab:headline_full}, excluded-cell CIs are iid. Grey ticks: the three secondary estimators (score matching, embedding+score matching, Crump-trimmed AIPW). The cells excluded from the main panel by the Section~\ref{sec:selection} criteria (Llama-2-13b-chat plus five small-$n_T$ cells) sit at visibly wider AIPW confidence intervals.}
  \label{fig:forest_full}
\end{figure*}

The main text restricts attention to the 13 main-panel cells, where the AIPW estimator's nuisance fit and IPW reweighting meet the inclusion criterion ($n_T \geq 500$, $\mathrm{ESS}_T \geq 100$, AIPW 95\% CI width $\leq 12$~pp; see Section~\ref{sec:selection}). For completeness this appendix reports the full 19-cell panel: the six excluded cells listed in Table~\ref{tab:excluded} are added back to a single all-cells table (Table~\ref{tab:headline_full}) and forest plot (Figure~\ref{fig:forest_full}). Read the excluded cells with the explicit caveats in Table~\ref{tab:excluded}.

\begin{table*}[!ht]
\small
\centering
\begin{tabular}{lrrrrl}
\toprule
Cell & $n_T$ & $\mathrm{ESS}_T$ & CI (pp) & AIPW (pp) & Binding constraint \\
\midrule
Llama-2-13b-chat (LMSYS) & 1{,}608 & 698 & \textbf{23.2} & $-24.5$ & CI > 12 (narrow overlap, $P(T)=0.68$) \\
Claude-1 (LMSYS) & \textbf{375} & 79 & 56.6 & $-25.1$ & $n_T < 500$ \\
Claude-instant-1 (LMSYS) & \textbf{185} & 36 & 59.3 & $+0.3$ & $n_T < 500$ \\
GPT-3.5-turbo (LMSYS) & \textbf{196} & 45 & 37.3 & $+3.9$ & $n_T < 500$ \\
GPT-4 (LMSYS) & \textbf{184} & 58 & 21.6 & $-16.3$ & $n_T < 500$ \\
GPT-4.1-mini (WildChat-general) & \textbf{178} & 152 & 6.3 & $+0.7$ & $n_T < 500$ \\
\bottomrule
\end{tabular}
\caption{Cells excluded from the main panel and the diagnostic that failed. ``CI'' is the AIPW 95\% confidence interval width in percentage points. Bolded entries are the binding constraint; if multiple, the most binding is shown.}
\label{tab:excluded}
\end{table*}

\begin{table*}[!ht]
\small
\centering
\begin{tabular}{lrrrrrr}
\toprule
Cell & $n_{\text{risky}}$ & Score & Emb & AIPW (95\% CI) & AIPW-T & AUC \\
\midrule
\multicolumn{7}{l}{\emph{LMSYS-Chat-1M}} \\
Llama-13b (base) & 5{,}040 & $-4.1$ & $+1.5^{ns}$ & $+0.3\,[-1.2,+1.8]^{\,g\,ns}$ & $+0.1$ & 0.62 \\
Llama-2-13b-chat & 2{,}365 & $-22.4$ & $-14.5$ & $-24.5\,[-36.1,-12.9]$ & $-18.0$ & 0.90 \\
Vicuna-13b & 2{,}207 & $-19.2$ & $-1.6^{ns}$ & $-7.1\,[-12.7,-1.4]^{\,g}$ & $-10.0$ & 0.83 \\
Vicuna-33b & 7{,}754 & $-32.8$ & $-12.8$ & $-11.9\,[-16.0,-7.7]^{\,g}$ & $-10.4$ & 0.87 \\
Koala-13b & 2{,}431 & $-16.0$ & $-10.6$ & $-3.1\,[-6.4,+0.2]^{\,g\,ns}$ & $-3.5$ & 0.77 \\
GPT-3.5-turbo & 579 & $-10.7$ & $-25.0^{ns}$ & $+3.9\,[-14.7,+22.6]^{ns}$ & $-11.1$ & 0.86 \\
GPT-4 & 622 & $-5.4^{ns}$ & $-38.5$ & $-16.3\,[-27.1,-5.5]$ & $-9.9$ & 0.78 \\
Claude-1 & 759 & $-20.8$ & $-20.6^{ns}$ & $-25.1\,[-53.4,+3.2]^{ns}$ & $-12.6$ & 0.83 \\
Claude-instant-1 & 404 & $+0.0^{ns}$ & $+7.1^{ns}$ & $+0.3\,[-29.4,+29.9]^{ns}$ & $-7.6$ & 0.79 \\
\midrule
\multicolumn{7}{l}{\emph{WildChat, general traffic}} \\
GPT-4o & 5{,}822 & $-7.6$ & $-8.0$ & $-7.6\,[-12.6,-2.7]^{\,u}$ & $-7.0$ & 0.88 \\
GPT-4o-mini & 7{,}348 & $-13.1$ & $-1.5$ & $-1.0\,[-2.4,+0.4]^{\,u\,ns}$ & $+0.3$ & 0.89 \\
GPT-4.1-mini & 9{,}844 & $-4.5^{ns}$ & $+0.0$ & $+0.7\,[-2.4,+3.9]^{ns}$ & $+4.0$ & 0.65 \\
\midrule
\multicolumn{7}{l}{\emph{WildChat, risky-enriched oversample}} \\
GPT-3.5-turbo & 29{,}936 & $+6.1$ & $+7.0$ & $+8.2\,[-5.9,+22.2]^{\,u\,ns}$ & $+9.5$ & 0.94 \\
GPT-4o & 14{,}966 & $-6.4$ & $-2.1$ & $-2.9\,[-4.4,-1.5]^{\,u}$ & $-3.8$ & 0.90 \\
GPT-4o-mini & 19{,}231 & $-14.1$ & $-6.2$ & $-4.9\,[-11.5,+1.8]^{\,u\,ns}$ & $-4.8$ & 0.92 \\
GPT-4.1-mini & 20{,}690 & $-2.1$ & $-0.1^{ns}$ & $-0.2\,[-0.7,+0.3]^{\,u\,ns}$ & $-0.2$ & 0.95 \\
GPT-4 (0314) & 21{,}400 & $-14.7$ & $-30.6$ & $-17.4\,[-28.3,-6.5]^{\,u}$ & $-14.8$ & 0.89 \\
GPT-4-Turbo (1106) & 36{,}410 & $-15.3$ & $-11.9$ & $-13.4\,[-19.7,-7.1]^{\,u}$ & $-14.4$ & 0.90 \\
GPT-4-Turbo (0125) & 24{,}769 & $-18.9$ & $-3.8$ & $-9.8\,[-14.8,-4.9]^{\,u}$ & $-10.8$ & 0.92 \\
\bottomrule
\end{tabular}
\caption{RAW benchmark results on the full 19-cell analysis panel (the numeric companion to Figure~\ref{fig:forest} in the main text; six excluded cells with caveats summarized in Table~\ref{tab:excluded}). ATT is the turn-3 re-engagement difference in percentage points. Score and Emb are the score-matching and embedding+score-matching ATT estimates (pp) of Table~\ref{tab:estimators}; AIPW carries a 95\% CI under the default inference regime: $^{u}$ marks hashed-IP user-clustered standard errors (WildChat main-panel cells), $^{g}$ marks near-duplicate-prompt-group clustered standard errors (LMSYS main-panel cells; the LMSYS release has no user identifier), and unmarked intervals, on the six excluded cells only, are iid. $^{ns}$ marks CIs that cross zero (on Score/Emb, pairwise CIs). AIPW-T is Crump-trimmed AIPW; AUC is propensity-model AUC$(T,X)$, higher meaning narrower overlap. Appendix~\ref{app:dependence} reports all three inference regimes side by side.}
\label{tab:headline_full}
\end{table*}

\paragraph{Reading the all-cells forest.} In Figure~\ref{fig:forest_full}, the excluded cells (Llama-2-13b-chat LMSYS, Claude-1, Claude-instant-1, GPT-3.5-turbo LMSYS, GPT-4 LMSYS, GPT-4.1-mini WildChat-general) sit at either large absolute AIPW point estimates (Llama-2-13b-chat $-24.5$, Claude-1 $-25.1$, GPT-4 $-16.3$) or near zero with very wide CIs (Claude-instant-1, GPT-4.1-mini), and in either case fail the precision criterion. Two patterns emerge that are obscured in Figure~\ref{fig:forest}: (i)~the most aggressively safety-tuned chat models (Llama-2-13b-chat, Claude-1) carry the largest \emph{full-sample} AIPW magnitudes alongside CIs far beyond the precision criterion; the literally widest intervals belong to the small-$n_T$ Claude cells, whereas Llama-2-13b-chat's interval is $23$~pp wide despite the largest treated count among excluded cells ($n_T = 1{,}608$), the signature of outcome-model extrapolation under narrow overlap rather than small-sample noise; (ii)~Crump-trimmed AIPW pulls all the excluded cells back toward zero relative to full-sample AIPW (Llama-2-13b-chat $-24.5 \to -18.0$, Claude-1 $-25.1 \to -12.6$), confirming overlap-driven inflation in the full-sample estimator on these high-treatment-prevalence cells.

\section{Inference under within-user and within-prompt dependence}
\label{app:dependence}

Conversations in these corpora are not independent: the same user submits many conversations, often resubmitting near-identical prompts. In the largest cell, GPT-4-Turbo (1106) risky-enriched, $92.9\%$ of the $36{,}410$ conversations come from a hashed IP with more than one conversation in the cell (per-cell user concentration in Table~\ref{tab:provenance}). Treating conversations as independent therefore understates standard errors. Table~\ref{tab:cluster} compares three inference regimes on the main panel: iid influence-function standard errors (the naive conversation-level baseline), hashed-IP user clusters (available on the nine WildChat cells; the LMSYS release has no user identifier of any kind), and near-duplicate prompt-group clusters (embedding cosine $\geq 0.95$; computable on all cells, and the feasible dependence adjustment on LMSYS). Point estimates are identical across regimes; only the intervals change.

\begin{table*}[!ht]
\small
\centering
\begin{tabular}{lrrrrr}
\toprule
Cell & ATT & iid CI & User-clustered CI & Users & Prompt-clustered CI \\
\midrule
\multicolumn{6}{l}{\emph{LMSYS-Chat-1M (no user identifier in source)}} \\
Llama-13b (base) & $+0.3$ & $[-1.2,+1.8]$ & -- & -- & $[-1.2,+1.8]$ \\
Vicuna-13b & $-7.1$ & $[-12.5,-1.6]$ & -- & -- & $[-12.7,-1.4]$ \\
Vicuna-33b & $-11.9$ & $[-16.0,-7.7]$ & -- & -- & $[-16.0,-7.7]$ \\
Koala-13b & $-3.1$ & $[-6.2,-0.0]$ & -- & -- & $[-6.4,+0.2]$ \\
\midrule
\multicolumn{6}{l}{\emph{WildChat, general traffic}} \\
GPT-4o & $-7.6$ & $[-11.9,-3.4]$ & $[-12.6,-2.7]$ & 3{,}441 & $[-12.8,-2.5]$ \\
GPT-4o-mini & $-1.0$ & $[-2.2,+0.2]$ & $[-2.4,+0.4]$ & 1{,}796 & $[-2.5,+0.6]$ \\
\midrule
\multicolumn{6}{l}{\emph{WildChat, risky-enriched oversample}} \\
GPT-3.5-turbo & $+8.2$ & $[+6.0,+10.3]$ & $[-5.9,+22.2]$ & 10{,}130 & $[+0.3,+16.0]$ \\
GPT-4o & $-2.9$ & $[-4.1,-1.7]$ & $[-4.4,-1.5]$ & 10{,}399 & $[-4.9,-0.9]$ \\
GPT-4o-mini & $-4.9$ & $[-8.0,-1.7]$ & $[-11.5,+1.8]$ & 4{,}270 & $[-10.1,+0.4]$ \\
GPT-4.1-mini & $-0.2$ & $[-0.6,+0.2]$ & $[-0.7,+0.3]$ & 1{,}017 & $[-0.7,+0.4]$ \\
GPT-4 (0314) & $-17.4$ & $[-21.5,-13.3]$ & $[-28.3,-6.5]$ & 3{,}929 & $[-24.9,-9.9]$ \\
GPT-4-Turbo (1106) & $-13.4$ & $[-14.6,-12.2]$ & $[-19.7,-7.1]$ & 4{,}483 & $[-18.3,-8.5]$ \\
GPT-4-Turbo (0125) & $-9.8$ & $[-12.0,-7.6]$ & $[-14.8,-4.9]$ & 3{,}676 & $[-14.1,-5.5]$ \\
\bottomrule
\end{tabular}
\caption{ATT (pp) with 95\% confidence intervals under three inference regimes, 13-cell main panel. User clusters are hashed IPs from the WildChat source release, re-attached by conversation hash ($100\%$ join, zero nulls, in every cell); Users counts distinct hashed IPs in the analysis pool. Prompt clusters are near-duplicate first-user-message groups at embedding cosine $\geq 0.95$. The paper's default regime (Table~\ref{tab:headline_full}) is user clusters where available and prompt clusters on the four LMSYS cells. Under iid inference the partition is 9 significant negative, 3 null, and 1 significant positive; under the default clustered regime it is 7 significant negative and 6 statistically indistinguishable from zero: GPT-4o-mini risky-enriched and the positive GPT-3.5-turbo cell lose $95\%$ significance under user clustering (the latter with $6.55\times$ standard-error inflation), and Koala-13b, already borderline (iid upper bound $-0.0$), loses it under prompt clustering. The dated GPT-4 line keeps significance in every regime.}
\label{tab:cluster}
\end{table*}

\paragraph{Calendar time, sampling parameters, and matching-design variants.} Table~\ref{tab:robustness} collects the robustness runs on the three largest WildChat-risky cells (all AIPW rows user-clustered). Adding conversation date to the covariate set moves the ATT by at most $0.13$ pp and the propensity AUC by less than $0.001$, even though the overall refusal rate drifts substantially within collection windows (on GPT-4-Turbo (1106), from $8.4\%$ in November 2023 to $25.4\%$ in March 2024, while control-arm re-engagement stays in a flat $0.36$--$0.41$ band): the drift is absorbed by the prompt-content and risk covariates already in $X$. Adding temperature and top-$p$ to $X$, or restricting to modal-parameter conversations, is inert because the parameters are constant in $99.7\%$ or more of these cells' conversations. On the matching side, restricting to same-calendar-month pairs moves estimates by at most $1.4$ pp, mainly because $85$--$96\%$ of baseline pairs were already same-month. Excluding same-IP pairs changes the estimand to a cross-user comparison; each refused prompt is re-matched to its best cross-user control rather than dropped, which roughly halves the pair count instead of removing the $77$--$84\%$ same-IP share outright (mean cosine falls to ${\approx}0.90$--$0.93$, so magnitudes are not directly comparable). The negative cells become at least as negative, while the positive cell collapses to $+1.9$ $[-3.5,+7.3]$, locating its effect in within-user persistence. Restricting to near-verbatim pairs ($\cos \geq 0.99$) leaves the GPT-4 cells essentially unchanged. Finally, splitting treated conversations at the median first-response length shows both terse and long refusals significantly costly on GPT-4-Turbo (1106), with terser refusals costlier; in the positive cell, whose median refusal runs $62$ characters of boilerplate, the positive point estimate is carried entirely by the terse stratum, coherent with the empty-send re-roll pattern of Appendix~\ref{app:taxonomy}.

\begin{table*}[!ht]
\small
\centering
\begin{tabular}{lccc}
\toprule
Variant & GPT-4-Turbo (1106) & GPT-4-Turbo (0125) & GPT-3.5-turbo (positive) \\
\midrule
AIPW, default & $-13.4\,[-19.7,-7.1]$ & $-9.8\,[-14.8,-4.9]$ & $+8.2\,[-5.9,+22.2]$ \\
\quad $+$ calendar time in $X$ & $-13.3\,[-19.6,-7.0]$ & $-9.9\,[-14.9,-5.0]$ & $+8.3\,[-5.8,+22.3]$ \\
\quad $+$ temperature, top-$p$ in $X$ & $-13.4\,[-19.7,-7.1]$ & $-9.8\,[-14.8,-4.9]$ & $+8.2\,[-5.9,+22.2]$ \\
\quad modal-parameter conversations only & $-13.3\,[-19.4,-7.1]$ & -- & -- \\
\quad treated: terse refusals only & $-16.6\,[-25.2,-8.0]$ & -- & $+14.3\,[-1.6,+30.2]$ \\
\quad treated: long refusals only & $-9.6\,[-14.6,-4.5]$ & -- & $-0.3\,[-12.0,+11.4]$ \\
\midrule
Emb.\ matching, baseline & $-11.9\,[-13.9,-10.0]$ & $-3.8\,[-7.3,-0.3]$ & $+7.0\,[+3.1,+11.0]$ \\
\quad same-calendar-month pairs & $-11.6\,[-13.5,-9.6]$ & $-2.6\,[-6.0,+0.8]$ & $+8.4\,[+4.4,+12.3]$ \\
\quad cross-user pairs only & $-27.6\,[-31.6,-23.5]$ & $-10.1\,[-17.1,-3.1]$ & $+1.9\,[-3.5,+7.3]$ \\
\quad near-verbatim pairs ($\cos \geq 0.99$) & $-11.7\,[-13.9,-9.6]$ & $-3.3\,[-7.2,+0.6]$ & $+4.1\,[-0.6,+8.9]$ \\
\bottomrule
\end{tabular}
\caption{Robustness variants on the three largest WildChat-risky cells, ATT in pp with 95\% CIs. AIPW rows use hashed-IP user-clustered standard errors; matching rows use pairwise CIs. Refusal-length strata split treated conversations at the treated-sample median first-response character length (controls untouched; medians $349$ characters on GPT-4-Turbo (1106) and $62$ on GPT-3.5-turbo). Cross-user matching excludes same-IP pairs, which are $77$--$84\%$ of baseline pairs, re-matching each refused prompt to its best cross-user control; its magnitudes are not directly comparable to baseline because pair composition loosens (mean cosine falls to ${\approx}0.90$--$0.93$). A dash marks variants not run for that cell.}
\label{tab:robustness}
\end{table*}

\section{Moderation-decile decomposition}
\label{app:decile}

\begin{figure*}[!htb]
  \centering
  \includegraphics[width=0.96\textwidth]{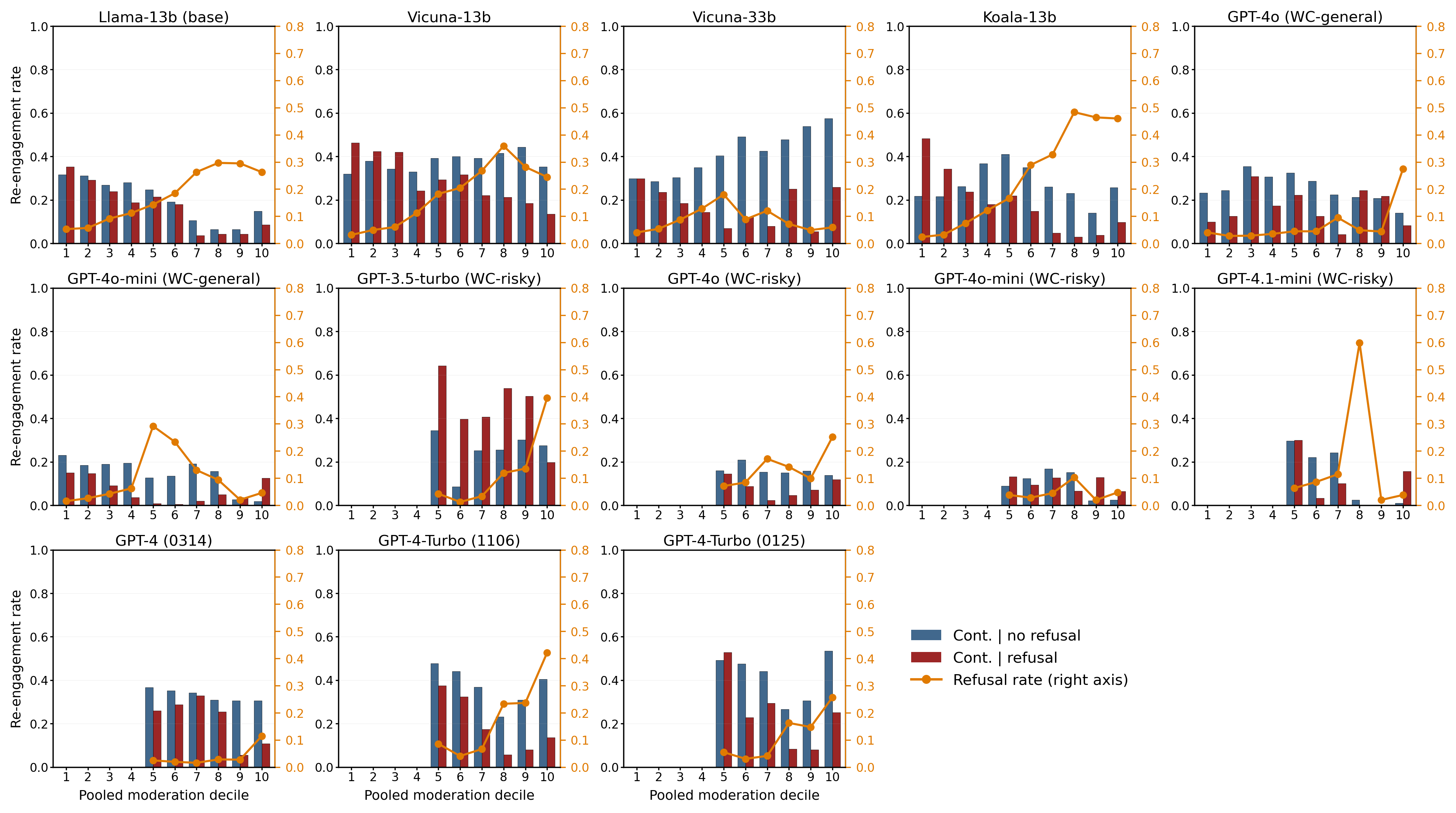}
  \caption{Refusal rate (orange line, right axis) and user re-engagement rate by treatment status (red = refused, blue = non-refused, left axis) at each pooled-quantile moderation decile, for the 13 main-panel cells. D1 (safest) and D10 (riskiest) refer to the same absolute \texttt{max\_concern\_score} band across every subplot. Risky-enriched cells contain only $X_s > 0.01$ conversations by construction, leaving D1--D4 empty there.}
  \label{fig:deciles_main}
\end{figure*}

Figure~\ref{fig:deciles_main} shows the full visual decile pattern for the 13 main-panel cells. Deciles are pooled-quantile bands over the panel-wide score distribution; each subplot is computed on that cell's full conversation pool, so LMSYS and WildChat-general subplots populate all ten deciles while risky-enriched cells contain only $X_s > 0.01$ conversations by construction.

Refusal rates rise monotonically with risk decile on most safety-tuned models, peaking at pooled-D10 levels of $85\%$ on Claude-instant-1 and $69\%$ on Llama-2-13b-chat among the excluded cells, and $46\%$ on Koala-13b, $42\%$ on GPT-4-Turbo (1106), and $39\%$ on GPT-3.5-turbo risky-enriched within the main panel. Within the GPT-4 line at the shared D10 ($X_s \geq 0.571$), GPT-4 (0314) refuses $11\%$, GPT-4-Turbo (1106) refuses $42\%$, and GPT-4-Turbo (0125) refuses $26\%$: refusal escalates after the March 2023 checkpoint and partially reverses by January 2024, while refusal aversion shrinks monotonically across the same releases (Figure~\ref{fig:forest}).

\section{Unadjusted vs.\ adjusted, full panel}
\label{app:adjusted}

\begin{figure*}[!htb]
  \centering
  \includegraphics[width=0.92\textwidth]{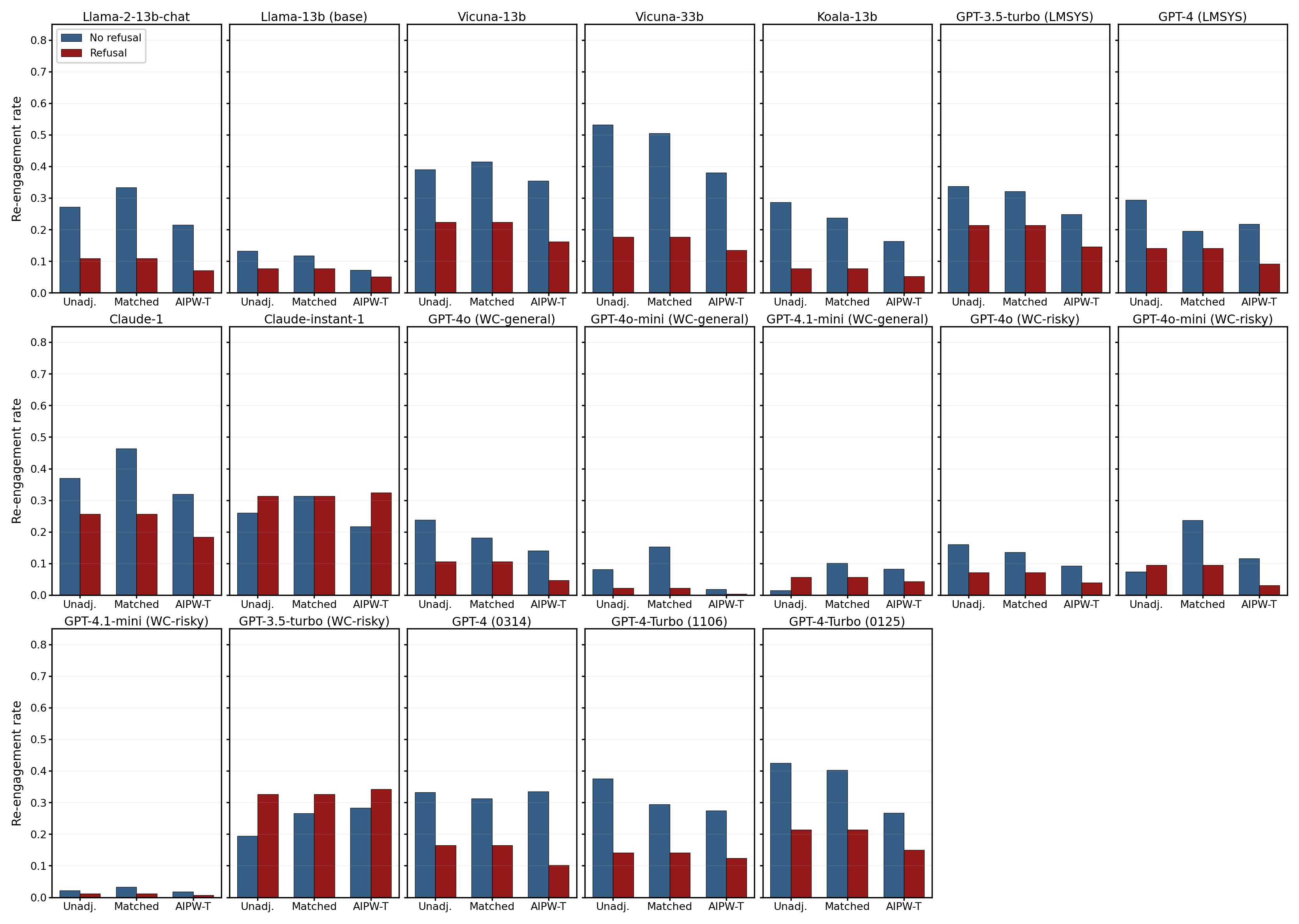}
  \caption{Re-engagement rates conditional on refusal status under three adjustments, for all 19 cells. Each panel shows one cell. Within each panel, two points are plotted per adjustment column (six per panel): the user re-engagement rate among \emph{refused} sessions (red marker) and among \emph{non-refused} sessions (blue marker); the gap between them is the ATT estimate under that adjustment. Adjustments: ``Unadj.'' uses no adjustment (the naive correlation in the risky pool); ``Matched'' uses score-caliper matching ($|\Delta X_s|\leq 0.1$); ``AIPW-T'' uses AIPW with Crump trimming ($\hat e \in [0.1, 0.9]$). The gap between the refused (red) and non-refused (blue) markers shrinks from the unadjusted column to the AIPW-T column on most cells; this shrinkage is the prompt-content-bias correction: refusals concentrate in risky prompts whose users have lower baseline continuation rates independent of the response.}
  \label{fig:adjusted_full}
\end{figure*}

Section~\ref{sec:headline} reports cells where naive correlations overstate refusal aversion. Figure~\ref{fig:adjusted_full} shows re-engagement rates conditional on refusal status, under three adjustments, for the full 19-cell panel.

\paragraph{Reading the adjusted figure.} The headline pattern in Figure~\ref{fig:adjusted_full} is that the unadjusted and AIPW-T markers diverge most on safety-tuned chat models (Vicuna-33b, Koala-13b, the dated GPT-4 line), where prompt-content composition strongly differs between refused and non-refused sessions. Three anomalies in the figure deserve attention. \emph{(a)~Llama-2-13b-chat and Claude-1} have full-sample AIPW estimates that are \emph{more} negative than the unadjusted gap ($-25$ vs.\ unadjusted $-16$ on Llama-2-13b-chat; $-25$ vs.\ unadjusted $-11$ on Claude-1), while Crump-trimmed AIPW returns to near the unadjusted value. This is the pathology behind the CI-width exclusion criterion (Section~\ref{sec:selection}): on cells where $P(T\mid \text{risky})$ is high ($0.68$ on Llama-2-13b-chat, $0.49$ on Claude-1), full-sample AIPW relies on outcome-model extrapolation into regions of poor overlap and over-corrects. \emph{(b)~GPT-4o-mini WildChat-risky} flips sign across the adjustment: unadjusted $+2$ pp, AIPW-T $-4.8$ pp, the only main-panel flip to an estimate away from zero. \emph{(c)~GPT-3.5-turbo WildChat-risky} is the panel's only cell with a positive ATT point estimate throughout: unadjusted $+13$ pp, AIPW-T $+9.5$ pp; under user-clustered inference its interval spans zero (Appendix~\ref{app:dependence}). We discuss its user-population reading in Section~\ref{sec:headline} and Appendix~\ref{app:taxonomy}.

\section{Cross-estimator agreement}
\label{app:cross_estimator}

\begin{figure*}[htbp]
  \centering
  \includegraphics[width=0.92\textwidth]{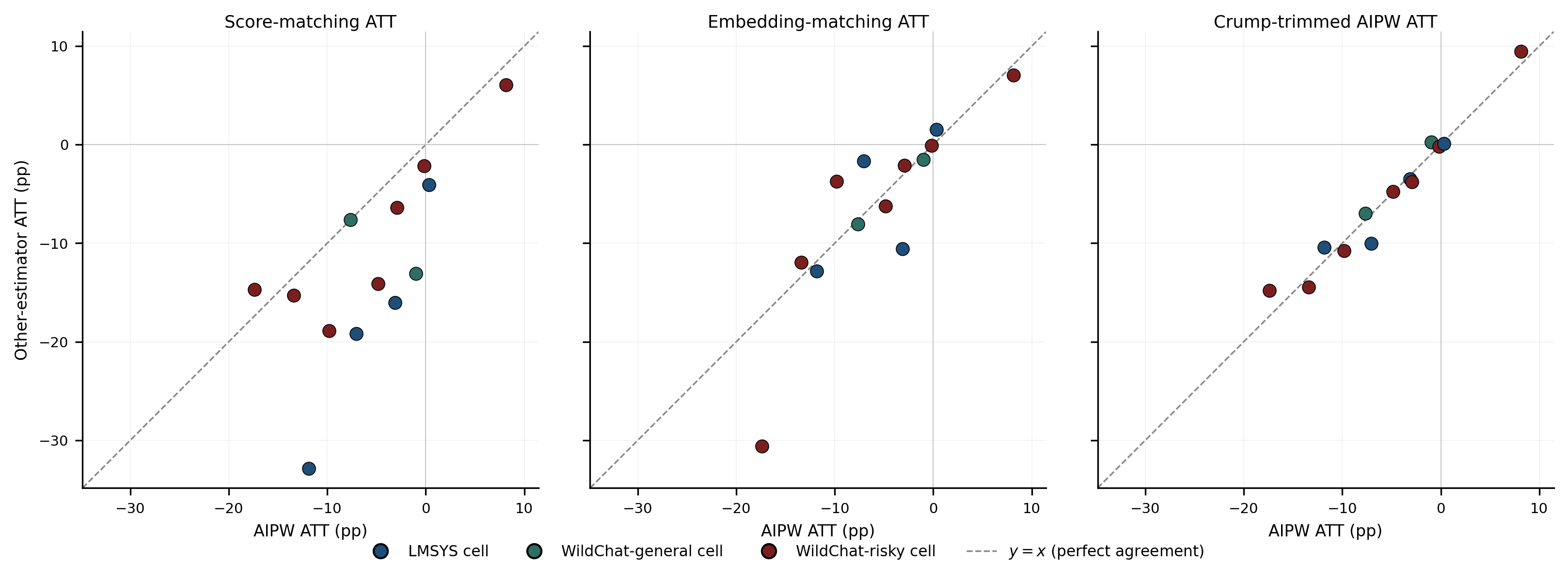}
  \caption{Cross-estimator agreement on the 13-cell main panel. $x$-axis: AIPW point estimate (pp); $y$-axis: secondary-estimator point estimate (pp). Each panel pairs AIPW against one of the three secondary estimators (Score matching, Embedding+score matching, Crump-trimmed AIPW). Each point is one cell, coloured by pool (blue = LMSYS, teal = WildChat-general, dark red = WildChat-risky). Dashed diagonal: $y = x$ (perfect agreement with AIPW). Points falling on the diagonal indicate the two estimators give the same ATT for that cell; departures from the diagonal indicate disagreement, with the direction of disagreement signalling which nuisance assumption fails.}
  \label{fig:cross_estimator}
\end{figure*}

\paragraph{Reading cross-estimator agreement.} In Figure~\ref{fig:cross_estimator}, the closest agreement is with Crump-trimmed AIPW (max gap $3.0$~pp on Vicuna-13b, computed from unrounded estimates); the points cluster tightly on the diagonal, indicating that AIPW's full-sample estimate and its overlap-restricted version converge on every main-panel cell. Embedding+score matching's largest gap ($13.2$~pp on GPT-4~(0314)) is driven by the small embedding-matched pair count ($n_{\text{pairs}}=330$) identifying a local effect on a different sub-population of near-duplicate prompts. Score matching's max gap ($21.0$~pp on Vicuna-33b, unrounded) reflects the scalar-score caliper conditioning on less prompt-content variation than the embedding does, exactly the misspecification route Table~\ref{tab:estimators} flags. The overall pattern is symmetric around the diagonal: the AIPW headline is corroborated by the secondary estimators, and the size and direction of disagreements diagnose \emph{which} nuisance assumption is binding rather than \emph{whether} the headline is sound.

\section{Propensity overlap}
\label{app:overlap}

\begin{figure*}[!htb]
  \centering
  \includegraphics[width=0.92\textwidth]{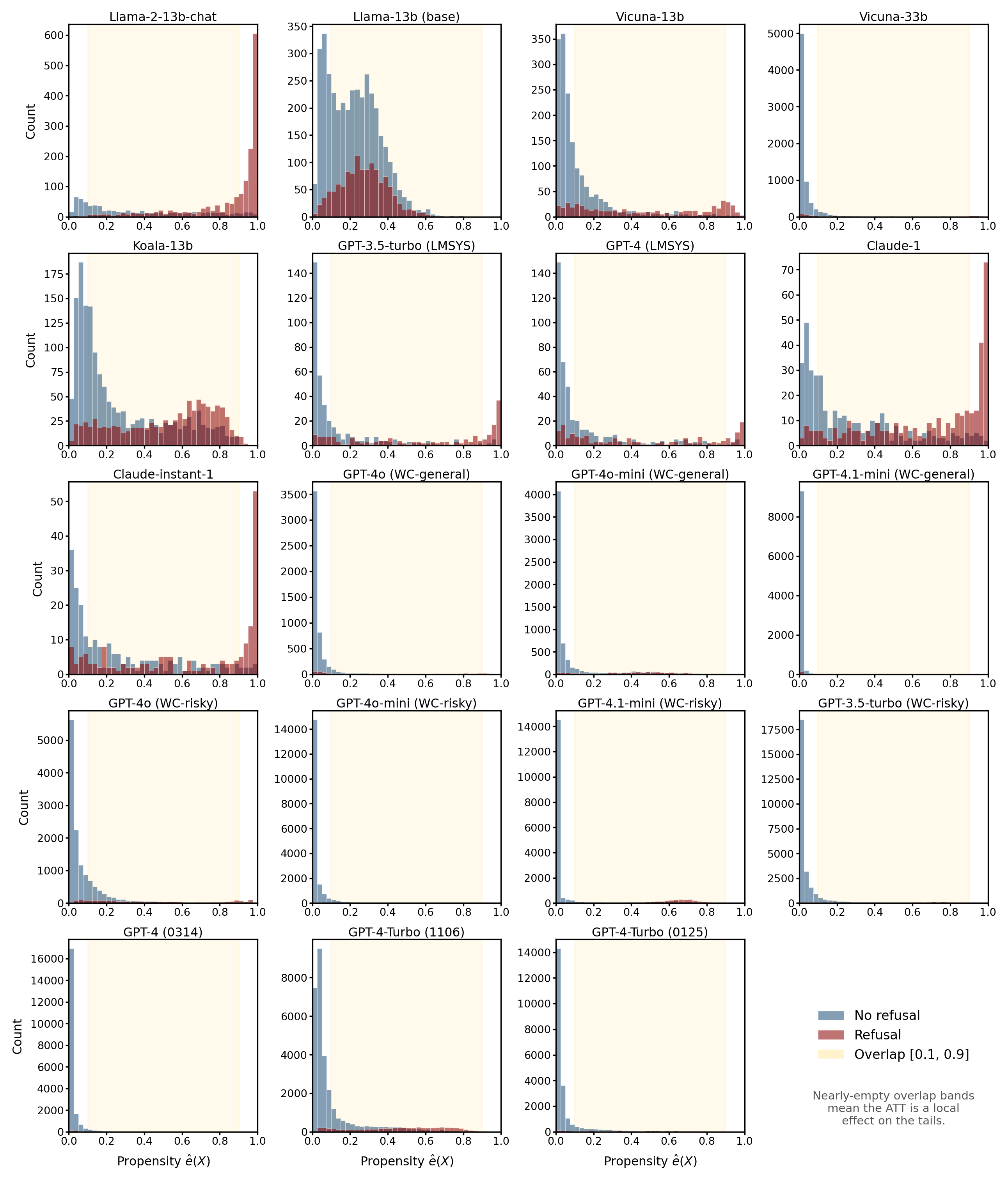}
  \caption{Cross-fit propensity $\hat e(X) = \Pr(T=1 \mid X)$ distribution by treatment status, all 19 cells. One subplot per cell. $x$-axis: propensity value $\hat e(X) \in [0,1]$; $y$-axis: count. Light histogram = non-refused conversations; dark histogram = refused conversations. The shaded vertical band $\hat e \in [0.1, 0.9]$ marks the Crump trim region: observations outside this band are excluded from the Crump-trimmed AIPW estimator. Wide overlap between the two histograms indicates the propensity model cannot perfectly separate refused from non-refused on $X$, which is what AIPW needs for the doubly-robust contrast to be supported by both arms.}
  \label{fig:propensity}
\end{figure*}

\paragraph{Reading propensity overlap.} Figure~\ref{fig:propensity} shows the asymmetry that motivates the trimmed estimator: most LMSYS cells have broad propensity distributions ($\mathrm{AUC} \approx 0.62$--$0.83$; Llama-13b base, for instance, is mixed across $[0.05,0.6]$ and supports full-sample AIPW without trimming), whereas the WildChat GPT-API cells (except GPT-4.1-mini general, AUC $0.65$) and Vicuna-33b concentrate both arms' mass near the extremes ($\mathrm{AUC} \geq 0.87$), where full-sample AIPW extrapolates into propensity-thin regions and the Crump-trimmed estimator is the appropriate check.

\section{Rosenbaum $\Gamma$ sensitivity bounds}
\label{app:gamma}

\begin{figure*}[!htb]
  \centering
  \includegraphics[width=0.85\textwidth]{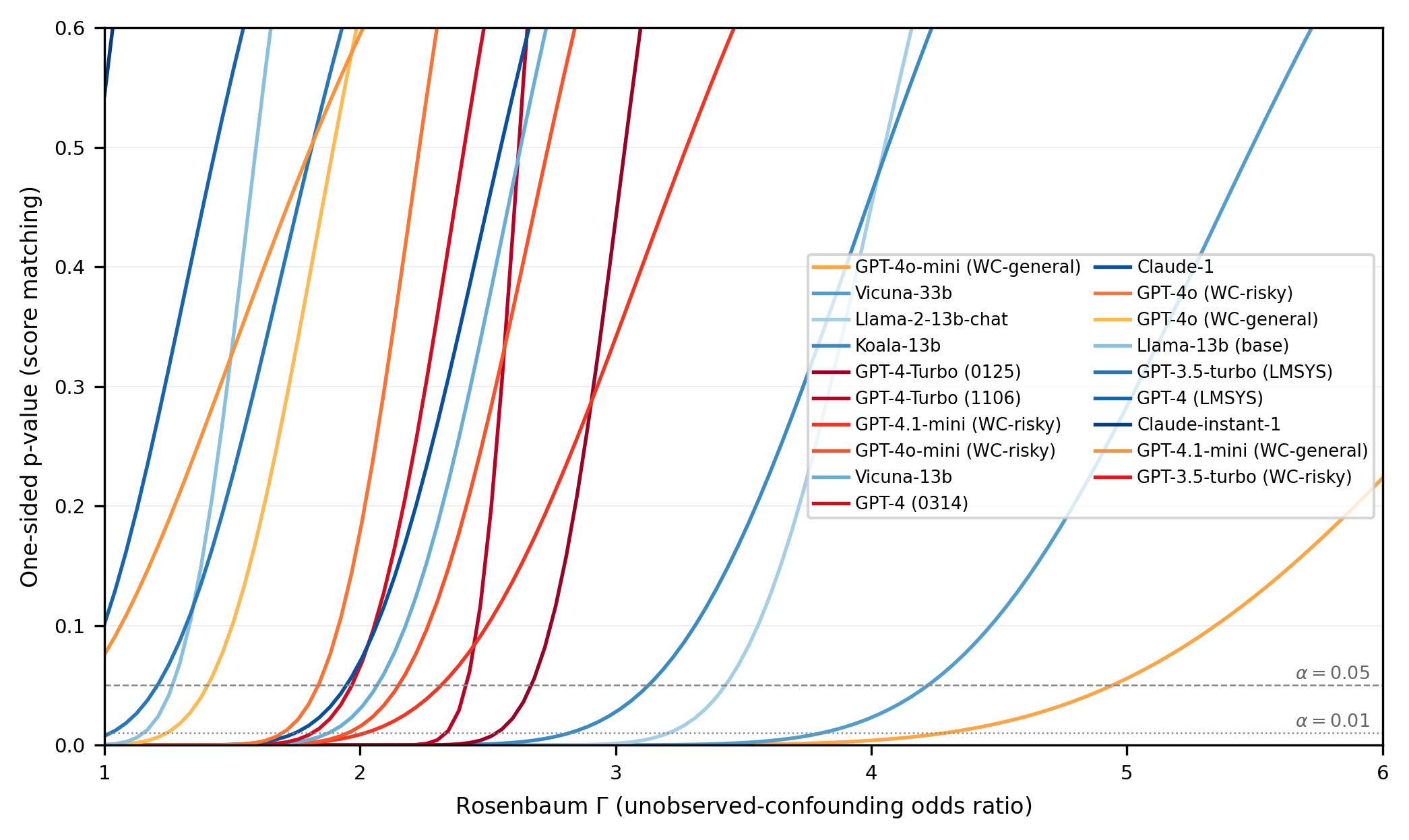}
  \caption{Rosenbaum $\Gamma$ sensitivity curves on the score-matching design, all 19 cells. $x$-axis: $\Gamma$, the assumed odds-ratio effect of a hidden confounder on refusal between matched pairs (left: $\Gamma=1$ means no hidden confounder; right: large $\Gamma$ means a strong hidden confounder). $y$-axis: upper-bound McNemar $p$-value at that $\Gamma$. Horizontal reference: $\alpha = 0.05$. The intersection of each cell's curve with $\alpha = 0.05$ gives $\Gamma_{0.05}$ for that cell, the smallest hidden-confounder strength that could push the matched-pair test to non-significance. Higher $\Gamma_{0.05}$ = more robust to unobserved confounding.}
  \label{fig:gamma}
\end{figure*}

\paragraph{Reading the $\Gamma$ curves.} In Figure~\ref{fig:gamma}, the substantively-strong threshold by the observational-studies convention is $\Gamma_{0.05} \geq 3$, meaning the hidden confounder would have to triple the refusal odds between matched pairs to overturn the result. Main-panel cells passing that bar under score matching: GPT-4o-mini WildChat-general ($\Gamma_{0.05}=4.94$), Vicuna-33b ($4.22$), Koala-13b ($3.13$), plus the excluded cell Llama-2-13b-chat ($3.43$). Three further main-panel cells pass under embedding matching: GPT-4~(0314) WildChat-risky ($6.34$, the panel maximum), GPT-4-Turbo (1106) WildChat-risky ($3.31$), GPT-4o WildChat-general ($3.21$), for six main-panel cells in total. Cells with $\Gamma_{0.05}$ near $1$ in the figure typically reflect a power problem (few discordant pairs after matching) rather than identification failure, visible as curves that cross $\alpha = 0.05$ almost immediately, at $\Gamma$ barely above $1$.

\section{What post-refusal re-engagement is: a turn-3 taxonomy}
\label{app:taxonomy}

The outcome $Y$ measures continued interaction, not satisfaction. To see what post-refusal re-engagement consists of, a \texttt{gpt-5.4-mini} judge classified the turn-3 user message of \emph{every} post-refusal re-engagement conversation in the main panel ($3{,}046$ turns; $T=1$, $Y=1$) and of $2{,}000$ control-arm re-engagement conversations sampled proportionally to cell size ($T=0$, $Y=1$), with $100\%$ judge completion. Categories: (A)~rephrase or retry of the same request; (B)~jailbreak or escalation attempt (persona play, DAN-style prompts, manipulation, threats); (C)~new topic; (D)~complaint about the response without retrying; (E)~thanks, closing, or small talk; (F)~other. For the control arm, category A reads ``repeats or refines their previous request,'' the natural follow-up, so a high control-arm A share is expected; the discriminating categories are B and D.

\begin{table}[t]
\footnotesize
\setlength{\tabcolsep}{4pt}
\centering
\begin{tabular}{lrrr}
\toprule
 & Treated & Control & Positive cell \\
Category & (pooled) & (pooled) & (treated) \\
\midrule
A: retry/rephrase & 58.3 & 82.8 & 60.8 \\
B: jailbreak/escalation & 19.5 & 6.5 & 17.3 \\
C: new topic & 9.8 & 7.8 & 8.2 \\
D: complaint & 8.1 & 1.0 & 10.7 \\
E: thanks/small talk & 2.8 & 1.1 & 2.2 \\
F: other & 1.5 & 0.8 & 0.8 \\
\midrule
A$+$B (persistence) & 77.9 & 89.3 & 78.1 \\
Empty turn-3 share & 13.0 & 0.7 & 41.0 \\
\bottomrule
\end{tabular}
\caption{Turn-3 taxonomy, percent. Category shares are among non-empty turn-3 messages ($2{,}651$ treated, $1{,}987$ control); the empty share is over all judged turns. ``Positive cell'' is GPT-3.5-turbo risky-enriched, treated arm. Control-arm B share within the positive cell is $4.6\%$, against $17.3\%$ after a refusal.}
\label{tab:taxonomy}
\end{table}

Three readings follow. First, $58\%$ of non-empty post-refusal turns retry or rephrase the refused request and a further $20\%$ escalate with jailbreak-style tactics, against a $6.5\%$ jailbreak share after compliance, and complaints run eight times the control arm's share. Post-refusal re-engagement is therefore mostly renewed pressure on the refused request, not retained engagement, which makes each negative cell's estimate a conservative floor: if part of measured re-engagement is adversarial retrying, the effect of refusal on \emph{satisfied} engagement is at least as negative as the measured ATT. Second, the positive cell's composition matches the panel: its induced re-engagement is persistence ($78\%$ retry or escalation; jailbreak share $17.3\%$ after refusal against $4.6\%$ after compliance). Third, $41\%$ of the positive cell's post-refusal turn-3 messages are \emph{empty sends}: zero-length user turns present as user messages in the source data (source-verified, not a redaction artifact; at most $5.5\%$ in every other cell, $1.3\%$ in the same cell's control arm). In a spot check of these conversations the model frequently produced the previously refused content at turn~4: the empty send operates as a re-roll of the sampler, and often succeeds, which coheres with the positive point estimate being carried by terse boilerplate refusals that are cheap to re-roll (Table~\ref{tab:robustness}).

\section{Treatment-label robustness and measurement error}
\label{app:label_robustness}

The treatment variable is an LLM-judge label with error. This appendix probes it two ways: re-estimating the panel under an independent judge, and bounding how much outcome-correlated mislabeling would be needed to overturn each cell's sign.

\paragraph{Independent-judge re-estimation.} Table~\ref{tab:wg} re-estimates the full main panel with the open-weights \texttt{wildguard} classifier \cite{han2024wildguard} as the sole treatment label, a panel-consistent single-judge relabeling (the headline label coalesces \texttt{gpt-5.4-mini} with \texttt{wildguard} fallback, so on cells where the OpenAI content filter rejected judge calls the two labelings share the fallback entries). The qualitative partition is judge-stable: the dated GPT-4 line gives $-13.8$/$-11.8$/$-6.8$ pp, all significant; Vicuna-33b, GPT-4o general, and GPT-4o risky-enriched keep sign and significance; and the positive cell gives $+8.9$ $[+6.8,+11.0]$. WildGuard intervals are iid and would widen under user clustering exactly as in Table~\ref{tab:cluster}, so read them for point-estimate stability rather than significance. Individual small-magnitude cells are not judge-stable: Vicuna-13b, Koala-13b, and GPT-4o-mini risky-enriched keep sign but lose significance, and the pretraining-only base cell moves from $+0.3$ (ns) to a small significant $-2.4$ $[-3.9,-0.8]$, so conclusions about any single small-magnitude cell should carry that uncertainty. The GPT-4o-mini general cell is the outlier in cross-judge disagreement (WildGuard relabels $82\%$ of its judge-treated conversations, Table~\ref{tab:iy}, and its treated count drops from $978$ to $339$), consistent with a soft-refusal-heavy response style; its estimate is null under both labels.

\begin{table}[t]
\footnotesize
\setlength{\tabcolsep}{1.8pt}
\centering
\begin{tabular}{lrrr}
\toprule
Cell & Coal. & WildGuard (95\% CI) & $n_T$ \\
\midrule
\multicolumn{4}{l}{\emph{LMSYS-Chat-1M}} \\
Llama-13b (base) & $+0.3^{ns}$ & $-2.4\,[-3.9,-0.8]$ & 1{,}969 \\
Vicuna-13b & $-7.1$ & $-1.8\,[-7.7,+4.0]^{ns}$ & 613 \\
Vicuna-33b & $-11.9$ & $-13.0\,[-18.3,-7.7]$ & 593 \\
Koala-13b & $-3.1^{ns}$ & $-2.0\,[-5.0,+0.9]^{ns}$ & 1{,}105 \\
\midrule
\multicolumn{4}{l}{\emph{WildChat, general traffic}} \\
GPT-4o & $-7.6$ & $-5.4\,[-7.9,-2.9]$ & 731 \\
GPT-4o-mini & $-1.0^{ns}$ & $+2.2\,[-1.4,+5.7]^{ns}$ & 339 \\
\midrule
\multicolumn{4}{l}{\emph{WildChat, risky-enriched}} \\
GPT-3.5-turbo & $+8.2^{ns}$ & $+8.9\,[+6.8,+11.0]$ & 2{,}673 \\
GPT-4o & $-2.9$ & $-2.1\,[-3.3,-0.9]$ & 2{,}019 \\
GPT-4o-mini & $-4.9^{ns}$ & $-2.8\,[-6.2,+0.5]^{ns}$ & 856 \\
GPT-4.1-mini & $-0.2^{ns}$ & $-0.0\,[-0.4,+0.4]^{ns}$ & 3{,}812 \\
GPT-4 (0314) & $-17.4$ & $-13.8\,[-18.7,-8.8]$ & 919 \\
GPT-4-Turbo (1106) & $-13.4$ & $-11.8\,[-13.1,-10.6]$ & 6{,}145 \\
GPT-4-Turbo (0125) & $-9.8$ & $-6.8\,[-9.2,-4.5]$ & 2{,}607 \\
\bottomrule
\end{tabular}
\caption{Main-panel ATT (pp) under the WildGuard-only treatment label. Coal.\ is the headline coalesced-label estimate, with $^{ns}$ following the default clustered regime of Table~\ref{tab:headline_full}; WildGuard intervals are iid influence-function CIs from the single-judge run. $n_T$ is the WildGuard-labeled treated count; it differs from the coalesced label's treated count by $-639$ to $+818$ across cells.}
\label{tab:wg}
\end{table}

\paragraph{Differential measurement-error sensitivity.} Following the sensitivity logic of \citet{imai2010measurement}, we ask how much \emph{outcome-correlated} mislabeling would be needed to overturn each cell's sign. Write the ATT as $\tau = m_1 - \mu_0$, where $m_1$ is the observed treated-arm re-engagement rate and $\mu_0$ the AIPW counterfactual. Two error channels exist. In the \emph{false-positive} channel, a fraction $\varepsilon$ of labeled-treated conversations are truly compliance (the soft-refusal case: ``I cannot do X, but here is Y'' labeled as refusal); if their outcome mean is $m_c$, the true treated mean is $(m_1 - \varepsilon m_c)/(1-\varepsilon)$, and the worst case for a negative $\tau$ sets $m_c = 0$, giving the sign-flip threshold $\varepsilon^{*}_{\mathrm{FP}} = 1 - m_1/\mu_0$. In the \emph{false-negative} channel, a fraction of labeled controls are truly refusals; treating $\mu_0$ as a simple mixture (an approximation, since $\mu_0$ is propensity-weighted), the worst case sets their outcome mean to $1$ and gives $\varepsilon^{*}_{\mathrm{FN}} = (\mu_0 - m_1)/(1 - m_1)$. For the positive cell the adversarial directions reverse. As a consistency check, substituting each threshold back into its adjustment formula returns $\tau = 0$ to machine precision. Table~\ref{tab:iy} reports both thresholds next to their empirical anchors: the share of paper-treated conversations that the independent WildGuard judge labels compliance (the false-positive anchor) and the share of paper-control conversations it labels refusal (the false-negative anchor), computed on the subset of each analysis pool where both judges emit a label.

\begin{table}[t]
\footnotesize
\setlength{\tabcolsep}{2.1pt}
\centering
\begin{tabular}{lrrrrr}
\toprule
 & & \multicolumn{2}{c}{False positive} & \multicolumn{2}{c}{False negative} \\
Cell & ATT & $\varepsilon^{*}_{\mathrm{FP}}$ & obs. & $\varepsilon^{*}_{\mathrm{FN}}$ & obs. \\
\midrule
\multicolumn{6}{l}{\emph{LMSYS-Chat-1M}} \\
Llama-13b (base) & $+0.3^{ns}$ & 0.4 & 24.2 & 4.3 & 28.3 \\
Vicuna-13b & $-7.1$ & 24.0 & 9.3 & 9.1 & 7.6 \\
Vicuna-33b & $-11.9$ & 40.2 & 18.4 & 14.4 & 1.5 \\
Koala-13b & $-3.1^{ns}$ & 28.9 & 4.3 & 3.4 & 16.2 \\
\midrule
\multicolumn{6}{l}{\emph{WildChat, general traffic}} \\
GPT-4o & $-7.6$ & 42.0 & 20.3 & 8.5 & 5.7 \\
GPT-4o-mini & $-1.0^{ns}$ & 30.4 & 82.0 & 1.0 & 2.6 \\
\midrule
\multicolumn{6}{l}{\emph{WildChat, risky-enriched}} \\
GPT-3.5-turbo & $+8.2^{ns}$ & 10.8 & 4.9 & 25.0 & 1.1 \\
GPT-4o & $-2.9$ & 28.9 & 12.5 & 3.1 & 2.4 \\
GPT-4o-mini & $-4.9^{ns}$ & 33.7 & 17.3 & 5.4 & 2.1 \\
GPT-4.1-mini & $-0.2^{ns}$ & 12.6 & 16.0 & 0.2 & 38.5 \\
GPT-4 (0314) & $-17.4$ & 51.3 & 11.7 & 20.8 & 0.7 \\
GPT-4-Turbo (1106) & $-13.4$ & 48.7 & 6.4 & 15.6 & 2.0 \\
GPT-4-Turbo (0125) & $-9.8$ & 31.5 & 4.5 & 12.5 & 1.9 \\
\bottomrule
\end{tabular}
\caption{Sign-flip thresholds under worst-case differential label error versus observed cross-judge disagreement, all in percent. $\varepsilon^{*}_{\mathrm{FP}}$: fraction of labeled-treated conversations that would have to be outcome-correlated false positives (worst-case outcome mean) to drive the ATT to zero; obs.: the share of paper-treated conversations WildGuard labels compliance. $\varepsilon^{*}_{\mathrm{FN}}$: same for labeled controls; obs.: the share of paper-control conversations WildGuard labels refusal. Thresholds for statistically null cells ($^{ns}$, default regime) are mechanically small and not meaningful.}
\label{tab:iy}
\end{table}

Three conclusions. For the seven significant cells, the closed-form false-positive threshold runs $24$ to $51\%$, between $2.1$ and $7.6$ times the observed treated-label disagreement, and the worst case additionally assumes every mislabeled conversation never re-engages; realistic soft-refusal errors, whose users re-engage at interior rates, would need to be several times more common still. Because the closed form holds $\mu_0$ fixed, we validated it by actually flipping the worst-case fraction and re-running the full cross-fit pipeline on two cells: the realized sign-flip point comes earlier ($24.5\%$ against the closed form's $28.9\%$ on GPT-4o risky-enriched; $17.6\%$ against $31.5\%$ on GPT-4-Turbo (0125)), and still sits $2.0$ and $3.9$ times above the observed disagreement. The false-negative thresholds also exceed their observed anchors on all seven significant cells, but three sit closer (Vicuna-13b, GPT-4o on both pools, margins $1.2$ to $1.5\times$) under that channel's maximally adversarial assumption that every hidden refusal among controls is followed by re-engagement. For exactly this case the independent-judge re-estimation above is the decisive check: it resolves every observed disagreement in WildGuard's favor, realized outcome rates included, and the two GPT-4o cells keep sign and significance while Vicuna-13b keeps sign. Soft or partial refusals remain the concrete channel to monitor, and an operator adopting RAW can reproduce this table from two judges' labels alone.

\section{Datasheet for RAW}
\label{app:datasheet}

Following \citet{gebru2021datasheets}.

\paragraph{Motivation.} RAW was created to enable causal measurement of the effect of LLM refusals on user conversation continuation in naturalistic chat, across multiple deployed LLMs. It supports the method described in the main text.

\paragraph{Composition.} Each instance is a conversation from LMSYS-Chat-1M (direct-chat subset) or WildChat-4.8M-Full, keyed by \texttt{conversation\_hash} or \texttt{conversation\_id}, with derived fields attached: per-message refusal labels (binary; coalesced gpt-5.4-mini-with-wildguard-fallback as the headline, plus the panel-wide wildguard pass for cross-judge sensitivity), per-message moderation scores (OpenAI Moderation API, attached to the source datasets), and an L2-normalized mxbai-embed-large embedding of the first user message. Total: $354{,}127$ conversations across $19$ (pool, model) cells with $558{,}739$ assistant-message refusal labels. No raw conversation text is redistributed; only derived labels, moderation scores, embedding vectors, and join keys back to source.

\paragraph{Collection.} Source datasets downloaded from HuggingFace, both access-gated: LMSYS-Chat-1M under a custom dataset license agreement (research and commercial use, no third-party redistribution of the dataset), WildChat-4.8M-Full under ODC-BY 1.0. Labels and embeddings computed by the authors using the pipeline described in Section~\ref{sec:method}.

\paragraph{Uses.} Any analysis of user-side behavior conditioned on LLM responses in naturalistic chat. We do not recommend using the refusal labels as ground-truth for training refusal classifiers; the labels are research artifacts, not gold-standard annotations.

\paragraph{Distribution.} The derived artifacts are distributed as a HuggingFace dataset (\url{https://huggingface.co/datasets/pkireyev1/raw-refusal-aversion-in-the-wild}) with Zenodo DOI \href{https://doi.org/10.5281/zenodo.22073974}{10.5281/zenodo.22073974} as the permanent identifier; the pipeline repository at \url{https://github.com/Pashasan/raw-refusal-aversion-in-the-wild} links both. Licenses follow each source. WildChat-derived rows are released under ODC-BY~1.0 with the required attribution to WildChat. For LMSYS-Chat-1M, whose license prohibits redistribution of the dataset, we release only conversation-hash join keys and our derived fields, and users obtain the source text through the LMSYS access gate. The accompanying pipeline code is released under the MIT license.

\end{document}